%% file: paper_lmu.tex
\documentclass[]{fairmeta}
\pdfoutput=1

\usepackage{amsmath,amssymb}
\usepackage{booktabs}
\usepackage{graphicx}
\graphicspath{{./}{assets/}}
\usepackage{hyperref}
\usepackage{multirow}
\usepackage{natbib}
\usepackage{siunitx}
\usepackage{subcaption}
\usepackage{listings}  % PyTorch-style pseudocode listing (appendix)
\usepackage{environ}
\usepackage{flafter}  % prevent any float (even [!t]) from drifting above its source position — keeps title block at top of page 1

\input{new_commands.tex}

\newenvironment{ack}{\section*{Acknowledgments}}{}

\title{
  Edit the Bits, Diff the Codes: Bitwise Residual Editing for Visual Autoregressive Models
}
\author[*]{Shengqiang Zhang}
\author[*]{Ruotong Liao}
\author{Volker Tresp}
\author{Barbara Plank}
\author{Hinrich Schütze}
\affiliation{LMU Munich \& Munich Center for Machine Learning (MCML)}
\contribution[*]{Equal contribution.}

\makeatletter
\let\abstract\relax

\NewEnviron{abstract}{\global\edef\abstractlist{{\noexpand\color{metafg}\unexpanded\expandafter{\BODY}}}}
\makeatother

\input{chapters/abstract.tex}

\metadata[Correspondence]{\url{shengqiang@cis.lmu.de}}
\metadata[Code]{\url{https://github.com/Shengqiang-Zhang/BitResEdit}}

\begin{document}
\thispagestyle{firstheader}

\maketitle

\input{chapters/intro.tex}
\input{chapters/method.tex}
\input{chapters/exp.tex}

\input{chapters/related_work.tex}
\input{chapters/conclusion.tex}

\begin{ack}
This work is supported by the DAAD programme Konrad Zuse Schools of Excellence in Artificial Intelligence, 
sponsored by the Federal Ministry of Research, Technology and Space.
\end{ack}

\bibliography{literature}
\bibliographystyle{plainnat}

\newpage
\appendix
\input{chapters/implementation_details.tex}
\input{chapters/pseudocode.tex}
\input{chapters/limitations.tex}
\input{chapters/failure_cases.tex}
\input{chapters/llm_usage.tex}

\end{document}

%% file: new_commands.tex
\providecommand{\BitEdit}{\textsc{BitEdit}}
\providecommand{\ResEdit}{\textsc{ResEdit}}
\providecommand{\BitResEdit}{\textsc{BitResEdit}}

\definecolor{kcmt}{HTML}{518B8B}   % comments  (teal)
\definecolor{kfun}{HTML}{D93282}   % function / method calls (magenta)
\definecolor{kop}{HTML}{2C2CFF}    % operators + - * / (blue)

\lstdefinestyle{kaiming}{%
  language=Python,
  basicstyle=\ttfamily\small,
  keywordstyle=\color{black},                 % control flow stays plain black
  commentstyle=\color{kcmt},
  alsoletter=_,                               % treat snake_case as one token
  emph={fwd,start,to_input,bsq,bsq_enc,up,sigmoid,bernoulli,kl_clip,decode,
        process,bern_kl,where,zeros,zeros_like,ones_like,randn,randn_like,
        sample,clip},
  emphstyle=\color{kfun},
  morecomment=[l]{\#},
  literate=%
    {+}{{{\color{kop}+}}}1
    {*}{{{\color{kop}*}}}1
    {/}{{{\color{kop}/}}}1
    {-}{{{\color{kop}-}}}1,
  showstringspaces=false,
  keepspaces=true,
  columns=fullflexible,
  aboveskip=0pt, belowskip=0pt,
  xleftmargin=1pt,
}

\newcounter{kalgctr}
\newenvironment{kalg}[2]{%
  \par\addvspace{0.7\baselineskip}\refstepcounter{kalgctr}%
  \noindent\begin{minipage}{\linewidth}%
    \setlength{\parskip}{0pt}\setlength{\parindent}{0pt}%
    \hrule height 0.9pt\vspace{3pt}%
    {\bfseries Algorithm \thekalgctr}~#1\par
    \def\tmpnote{#2}\ifx\tmpnote\empty\else
      \vspace{1.5pt}{\footnotesize Note: #2}\par\fi
    \vspace{3pt}\hrule height 0.45pt\vspace{3pt}%
}{%
    \vspace{3pt}\hrule height 0.45pt%
  \end{minipage}\par\addvspace{0.7\baselineskip}%
}

%% file: chapters/abstract.tex
\begin{abstract}
Text-guided image editing with visual autoregressive (VAR) generators requires
controlling both what the model samples and where the sampled change is written
back into the image code. Existing VAR editors mainly operate on token streams,
features, or flat next-token logits, leaving two native structures of
bitwise-residual VAR models underused: the per-bit Bernoulli prediction head and
the additive multi-scale residual code field from which the image is assembled.
We propose \BitResEdit{}, a training-free editor for bitwise-residual VAR
generators such as \textsc{Infinity}. \BitEdit{} performs source-negative
guidance by tilting the post-CFG per-bit log-odds along a source--target
contrast computed on a shared edited prefix, then projects each update into a
closed-form Bernoulli-KL trust region around the clean CFG sampler.
\ResEdit{} converts the sampled bits into per-scale continuous-code residuals,
gates them with a localization mask, and re-injects them through the
generator's native sum-of-scales. Together they couple decision-time bit
guidance with combination-time code composition, so masked-out latent features
are preserved exactly by code arithmetic while localized, scale-aware edits
are applied inside the target region. On PIE-Bench with Infinity-2B,
\BitResEdit{} attains the strongest text alignment among same-backbone VAR
editors, improving CLIP on the edited region by $+1.07$ over the strongest
prior editor while keeping background preservation competitive with it.
Ablations
show \BitEdit{} and \ResEdit{} play complementary roles in target alignment
and background preservation.
\end{abstract}

%% file: chapters/intro.tex
% Helper macros for prompt-diff captions in the intro teaser figure.
% \promptchg{old}{new}: word substitution -- red old, green new with arrow.
% \promptadd{phrase}:   inserted phrase -- green with leading "+".
% \promptdel{phrase}:   deleted phrase -- red with strikethrough (requires
%                       \usepackage{ulem}; falls back to red color if absent).
\providecommand{\promptchg}[2]{{\color{red}#1}\,\ensuremath{\rightarrow}\,{\color{green!50!black}#2}}
\providecommand{\promptadd}[1]{{\color{green!50!black}\ensuremath{+\,}#1}}
\providecommand{\promptdel}[1]{{\color{red}#1}}

\begin{figure}[!ht]
  \centering
  % Each cell: source image (left) | edit image (right) and a prompt-diff caption.
  \newcommand{\introcell}[3]{%
    % #1 = key (filename stem), #2 = caption prefix, #3 = caption suffix with diff.
    \begin{minipage}[t]{0.325\linewidth}\centering
      \includegraphics[width=0.485\linewidth]{figures/intro_examples/#1_src.jpg}\hfill
      \includegraphics[width=0.485\linewidth]{figures/intro_examples/#1_edit.jpg}\\[2pt]
      {\scriptsize #2 #3\par}
    \end{minipage}%
  }
  \introcell{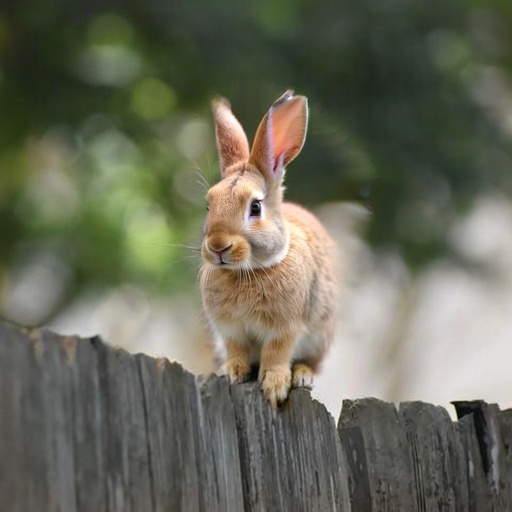}{a}{\promptchg{squirrel}{rabbit} sitting on top of a wooden fence}\hfill
  \introcell{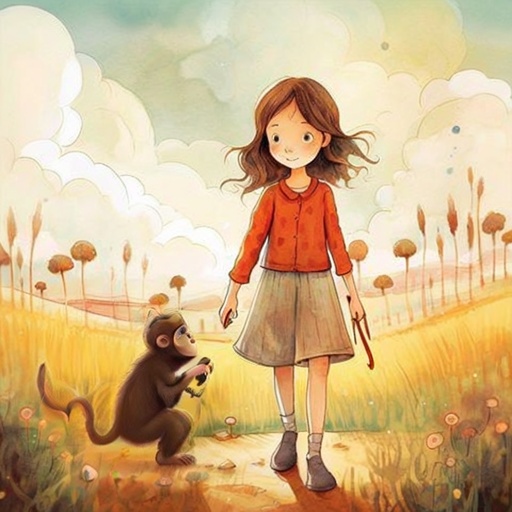}{a girl and her}{\promptchg{dog}{monkey} in a field}\hfill
  \introcell{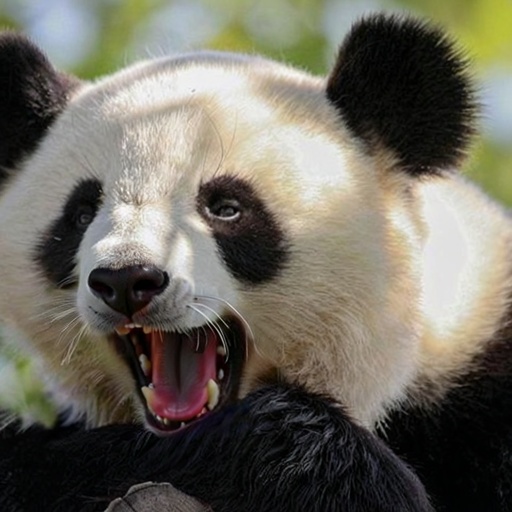}{a panda bear}{\promptchg{close}{open} his mouth}\\[6pt]
  \introcell{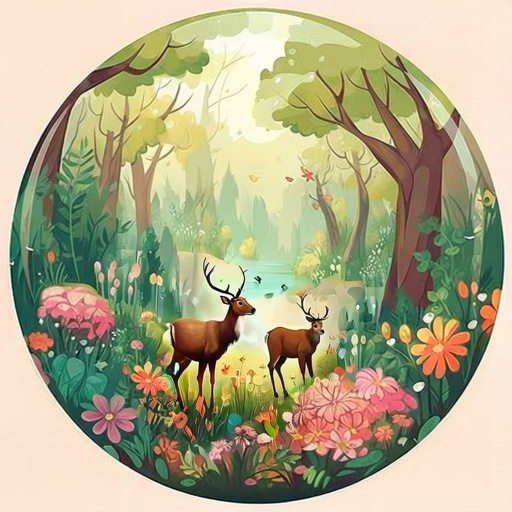}{a round painting of a forest with}{\promptadd{deer,} flowers, trees}\hfill
  \introcell{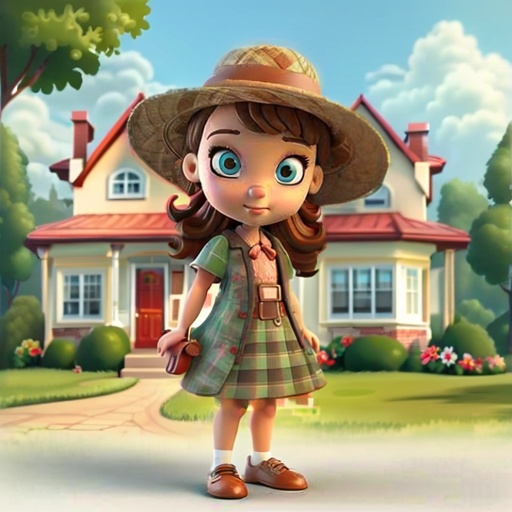}{a cartoon girl in a hat and dress}{\promptadd{standing in front of a beautiful house}}\hfill
  \introcell{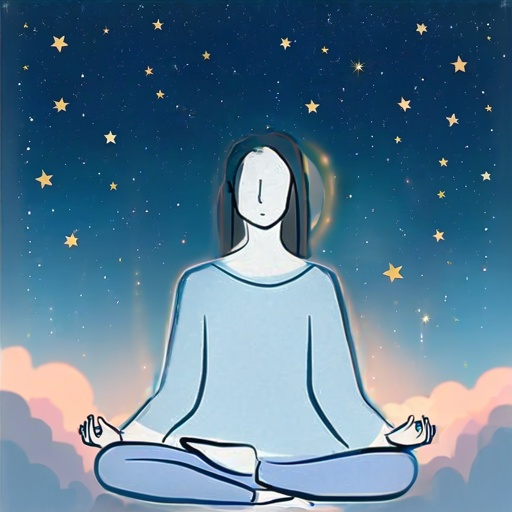}{illustration of a woman meditating in a yoga pose}{\promptadd{in the sky with stars}}
  \caption{Text-guided edits by \BitResEdit{}, our training-free editor for
  visual autoregressive models. Each pair shows the source (left) and edited
  (right) image; colored prompt diffs mark substitutions and insertions.
  \BitResEdit{} applies the requested change while preserving the rest of the
  image.}
  \label{fig:intro-teaser}
\end{figure}

\section{Introduction}
\label{sec:introduction}

Text-guided image editing modifies a user-specified region of a source image
according to a natural-language instruction---substituting objects, changing
attributes, or inserting content---while preserving unrelated content
(Fig.~\ref{fig:intro-teaser}).
Diffusion editors address this through inversion, guidance, attention or
feature control, and instruction tuning~\citep{brack2023ledits++,xu2024infedit,tang2024locinv,liu2025s2edit,bai2024edicho,sheynin2024emu,zhao2024ultraedit,wei2024omniedit,yu2025anyedit}.
Visual autoregressive (VAR)
generators~\citep{tian2024visual,han2025infinity} now reach high image quality
with lower latency, but expose a different mechanism: they sample
discrete multi-scale codes rather than denoise a continuous trajectory.
Diffusion-specific editing tools therefore do not transfer directly.
These differences are not only implementation details. In VAR models, an edit
decision at a coarse scale can constrain later fine-scale details, while a
late token change can still disturb local texture. A practical editor must
therefore decide where to change the autoregressive distribution and how to
write that change back into the image code. This motivates looking
inside the generator rather than adapting diffusion controls as black-box
perturbations.

We focus on \emph{bitwise-residual} VAR generators such as
\textsc{Infinity}~\citep{han2025infinity}, in which each token is a
$D$-bit code indexing one scale of a multi-scale residual code
field, and the prediction head emits independent per-bit Bernoulli
log-odds at every (scale, position, bit), instead of a $2^{D}$-way softmax
over the token vocabulary. Existing VAR editors mainly work at
the token or feature level: AREdit~\citep{wang2025aredit} recombines cached
source and target token streams under a mask, VAREdit~\citep{mao2025varedit}
learns scale-aligned reference injection, and Masked Logit
Nudging~\citep{elghoussani2026mln} softly biases VAR token logits using
source--target prompt differences. These methods improve VAR editing but
operate on discrete tokens, features, or token logits, not in the continuous
residual-code field from which the image is assembled; the per-bit
Bernoulli geometry likewise goes unused, and token-logit guidance lacks an
explicit per-bit bound on drift from a clean classifier-free guidance (CFG)
sampler.

This gap follows from two native properties of these generators.
Before sampling, the prediction head factorizes each token into independent
per-bit Bernoullis, so residual-code distributions admit a closed-form per-bit
KL bound. After sampling, the multi-scale residual codes combine into the final
VAE latent by a fixed linear sum. This sum makes the code space the natural
place to localize an edit: a mask-gated residual composes algebraically against
a cached source code and vanishes exactly outside the mask, so localization is
enforced by the generator's own arithmetic rather than attention maps or
token recombination. These facts suggest a split design with one component per
property: guide what the model samples with calibrated bit-level intervention,
then realize the sampled change as a mask-gated residual in code space.
Token-level editors and categorical logit nudging miss one or both properties,
while diffusion models expose neither.

Building on this observation, we propose \BitResEdit{}, a training-free
editor for bitwise-residual VAR generators. We assume access to a localization
mask, either provided by the benchmark or predicted by an external
grounding-and-segmentation pipeline. \BitEdit{} tilts the post-CFG log-odds at
every (scale, position, bit) along a source--target contrast computed on a
shared edited prefix, then projects the result into a per-bit Bernoulli-KL trust
region around the clean CFG sampler. This keeps each bit-level edit close to
the model's target-prompt behavior while adding source-aware pressure.
\ResEdit{} converts each scale's sampled bits into a continuous-code residual
against the cached source image, gates it by the localization mask, and
re-injects it through \textsc{Infinity}'s native sum-of-scales. At masked-out
positions the residual is exactly zero, so background preservation is anchored
in the generator's residual-code arithmetic up to the VAE decoder's boundary
effects. Coupled, \BitResEdit{} realizes calibrated bit-level decisions as
localized, scale-aware code-space edits.

\paragraph{Contributions.}
\textbf{(i)} We identify two algebraic properties that make bitwise-residual
VAR generators natively editable: per-bit Bernoulli factorization at the
prediction head, admitting a closed-form per-bit KL bound, and additive
multi-scale code-sum at latent assembly, admitting gated algebraic
composition against a cached source code. \textbf{(ii)}
We instantiate them as \BitEdit{}, a Bernoulli-KL trust-region source-negative
guidance computed on a shared edited prefix, and \ResEdit{}, a mask-gated
continuous-code residual re-injected through the native sum-of-scales.
\textbf{(iii)} On PIE-Bench, \BitResEdit{} attains the strongest text
alignment among same-backbone VAR editors: $+0.72$ whole-image and $+1.07$
edited-region CLIP over the strongest prior Infinity-2B editor, while
keeping background preservation competitive with it. Ablations show \BitEdit{} and \ResEdit{} play
complementary roles in target alignment and background preservation.

%% file: chapters/method.tex
\section{Bitwise Residual Editing}
\label{sec:method}

\BitResEdit{} edits in two steps. \BitEdit{} guides which bits are
sampled; \ResEdit{} controls where the sampled change is written into
the image code. We first review the structures both steps act on.

% Main method figure (TikZ skeleton)
% \input{chapters/fig_method.tex}

\subsection{Preliminaries: Bitwise residual visual autoregressive modeling}
\label{sec:preliminary}

\paragraph{Multi-scale residual generation.}
We build on \textsc{Infinity}~\citep{han2025infinity}, a bit-quantized VAR
generator. Given a text prompt $y$, \textsc{Infinity} produces an image's
continuous VAE code $z \in \mathbb{R}^{C \times H_z \times W_z}$ as a sum
of $K$ coarse-to-fine \emph{residual} fields,
\begin{equation}
\label{eq:sum_of_scales}
z = \sum_{k=1}^{K} \mathrm{up}_{k\to K}\!\left( c^{k} \right),
\quad c^{k} \in \mathbb{R}^{C \times H_k \times W_k},
\end{equation}
where $\mathrm{up}_{k\to K}$ bilinearly upsamples to $H_z \times W_z$.
Each scale uses a Binary Spherical Quantizer
(BSQ;~\citealp{zhao2024bsq,han2025infinity}). At every spatial
position $p \!\in\! \{1,\dots,P_k\}$ ($P_k = H_k W_k$), the tokenizer
produces a $D$-bit sign vector $b^{k}_{p} \!\in\! \{-1,+1\}^{D}$.
Following BSQ, we view the signed vector as a normalized code on the
unit hypersphere, $q^{k}_{p} = b^{k}_{p}/\sqrt{D}$, in the tokenizer
latent space, and write
$\mathrm{BSQ}\colon \{-1,+1\}^{D}\!\to\!\mathbb{R}^{C}$,
$\mathrm{BSQ}(b^{k}_{p}) = b^{k}_{p}/\sqrt{D}$, identifying $C\!=\!D$
for notational simplicity. Any learned projection between bit-feature
and decoder channels is absorbed into the tokenizer/decoder and
omitted from this notation. Thus
$c^{k}_{p} \!=\! \mathrm{BSQ}(b^{k}_{p})$ is the continuous code at
$(k,p)$, and the image is reconstructed as $\hat{x} = \mathcal{D}(z)$
through a fixed VAE decoder $\mathcal{D}$.

\paragraph{Per-bit Bernoulli head and CFG.}
At every scale $k$ the transformer outputs 2-class logits
$\ell^{k}(y) \!\in\! \mathbb{R}^{B \times P_k \times D \times 2}$, one
categorical distribution per (position, bit), conditioned on $y$ and the coarser
scales. We adopt the per-bit log-odds
$d^{k}(y) = \ell^{k}_{:,:,:,1}(y) - \ell^{k}_{:,:,:,0}(y)$ as our
working representation, since each bit is sampled from a Bernoulli
distribution $\mathrm{Bern}(\sigma(d^{k}_{p,j}(y)))$. Classifier-free
guidance (CFG;~\citealp{ho2022classifier}) is linear in $d$ and is
applied at the same visual prefix $\tilde c^{<k}$:
\begin{equation}
\label{eq:cfg}
d^{k}_{\mathrm{cfg}}(\tilde c^{<k})
= d^{k}(\varnothing \,\Vert\, \tilde c^{<k})
+ s\!\cdot\!\bigl(d^{k}(T \,\Vert\, \tilde c^{<k})
                  - d^{k}(\varnothing \,\Vert\, \tilde c^{<k})\bigr),
\end{equation}
with target prompt $T$, unconditional branch $\varnothing$ (a learned
unconditional text embedding in \textsc{Infinity}, not a literal empty
string), and CFG scale $s\!\geq\!1$. We will keep the prefix
$\tilde c^{<k}$ implicit in later equations when it is unambiguous.

\paragraph{Two intervention sites and editing notation.}
Eqs.~\eqref{eq:sum_of_scales}--\eqref{eq:cfg} expose two handles for
training-free editing: the per-bit log-odds $d^{k}$ \emph{before
sampling} (which bits are likely) and the per-scale code field $c^{k}$
\emph{after sampling} (where and how strongly a code change is
realized). Throughout, $S$ is the source prompt and $T$ the target.
Given a source image $x_{\mathrm{src}}$, the \emph{source pass} runs
the BSQ tokenizer on $x_{\mathrm{src}}$ and caches the resulting
per-scale source code fields $\{c^{k}_{\mathrm{src}}\}$; the tokenizer
itself is not prompt-conditioned. The \emph{edit pass} regenerates
scales $1{:}K$ to produce the edited code
$\tilde z = \sum_{k} \mathrm{up}_{k\to K}(\tilde c^{k})$, decoded to
the edited image $\tilde x = \mathcal{D}(\tilde z)$.

\begin{figure}[!t]
  \centering
  \includegraphics[width=\linewidth, trim=10 17 10 10, clip]{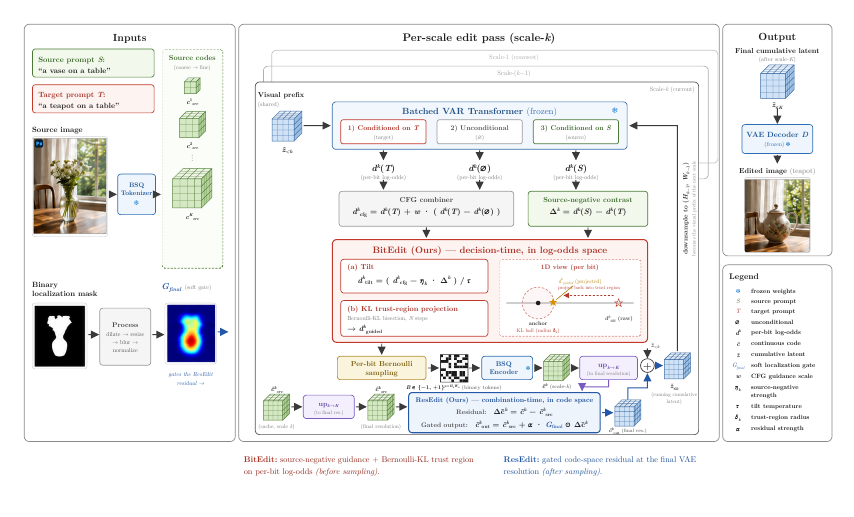}
  \caption{\textbf{\BitResEdit{} pipeline.} A source pass caches BSQ codes and
  the processed edit mask. Each edit scale applies \BitEdit{} to guide bit
  sampling and \ResEdit{} to write mask-gated code residuals before final decoding.}
  \label{fig:method}
\end{figure}

% --- 3.2 BitEdit -----------------------------------------------------------
\subsection{\BitEdit: Bernoulli trust-region source-negative guidance}
\label{sec:bitedit}

\BitEdit{} tilts the post-CFG Bernoulli log-odds along a per-bit
source--target contrast direction and projects the result onto a
Bernoulli-KL trust region around the post-CFG target. At each scale
$k$, a 3-branch forward returns
\begin{equation}
\label{eq:three_branch}
\bigl[\, d^{k}(T \,\Vert\, \tilde c^{<k}),\;
        d^{k}(\varnothing \,\Vert\, \tilde c^{<k}),\;
        d^{k}(S \,\Vert\, \tilde c^{<k}) \,\bigr],
\end{equation}
where the three log-odds fields are evaluated at the same edited
prefix $\tilde c^{<k}$ and differ only in the text condition. At the
first scale the visual prefix is empty and each branch starts from
its own text-derived start map. The
signed per-bit contrast
\begin{equation}
\label{eq:neg_direction}
\Delta^{k} \triangleq d^{k}(S \,\Vert\, \tilde c^{<k}) - d^{k}(T \,\Vert\, \tilde c^{<k})
\;\in\; \mathbb{R}^{B \times P_k \times D}
\end{equation}
encodes the target-relative source advantage at every bit. Anchoring
the contrast at $d^{k}(T \,\Vert\, \tilde c^{<k})$ rather than at
$d^{k}(\varnothing \,\Vert\, \tilde c^{<k})$ keeps directions shared
between source and target out of the guidance budget.

\paragraph{The \BitEdit{} update.}
Let $\tau\!>\!0$ be the sampling temperature,
$d^{k}_{\mathrm{cfg},\tau}\!\triangleq\!d^{k}_{\mathrm{cfg}}/\tau$ the
post-temperature CFG anchor, and factorize the per-bit guidance strength as
$\eta^{k}_{p,j}=\eta_{k}\,M^{k}_{p,j}$, with per-scale scalar $\eta_{k}\!\geq\!0$
(schedule below) and a spatial-bit mask $M^{k}_{p,j}\!\in\![0,1]$ that zeros
guidance at non-edit positions. Throughout we set $M^{k}$ to the binary
edit-region gate derived from the same localization mask as \ResEdit{}
(\S\ref{sec:resedit}), so guidance acts only inside the edit region;
$M^{k}\!\equiv\!1$ recovers unmasked guidance. The raw source-negative guided
log-odds is
\begin{equation}
\label{eq:bitedit_raw}
d^{k}_{\mathrm{raw},p,j}
\;=\; \bigl(d^{k}_{\mathrm{cfg},p,j} - \eta^{k}_{p,j}\, \Delta^{k}_{p,j}\bigr) / \tau,
\end{equation}
where dividing the full bracket by $\tau$ keeps $d^{k}_{\mathrm{raw}}$ and
$d^{k}_{\mathrm{cfg},\tau}$ on a common temperature scale, so their
Bernoulli-KL distance is directly comparable.

Large $\eta_{k}$ can push $d^{k}_{\mathrm{raw}}$ off-distribution, so we project
it onto a per-bit Bernoulli-KL ball of radius $\delta_{\mathrm{KL}}\!>\!0$
around $d^{k}_{\mathrm{cfg},\tau}$ by taking the largest mixing weight that
stays inside:
\begin{equation}
\label{eq:bitedit_proj}
d^{k}_{\mathrm{guided}}
\;=\; \lambda^{\star}\, d^{k}_{\mathrm{raw}} + (1-\lambda^{\star})\, d^{k}_{\mathrm{cfg},\tau},
\end{equation}
where
\begin{equation}
\label{eq:bitedit_alpha}
\lambda^{\star}
\;=\; \max\!\Bigl\{\,\lambda \!\in\! [0,1]
       \,\big|\,
       \mathrm{KL}\!\bigl(\mathrm{Bern}(\sigma(\lambda\,d^{k}_{\mathrm{raw}} + (1{-}\lambda)\,d^{k}_{\mathrm{cfg},\tau}))
       \,\big\Vert\,
       \mathrm{Bern}(\sigma(d^{k}_{\mathrm{cfg},\tau}))\bigr) \!\leq\! \delta_{\mathrm{KL}}\Bigr\}.
\end{equation}
The Bernoulli-KL is monotone along this segment, so $\lambda^{\star}$ is solved
per bit by bisection; bits
already feasible at $\lambda\!=\!1$ retain the full update. A secondary
element-wise clamp caps the deviation
$|d^{k}_{\mathrm{guided}}-d^{k}_{\mathrm{cfg},\tau}|$ as a safety rail for
outlier bits (value in Appendix~\ref{app:impl-details}). The clamped
$d^{k}_{\mathrm{guided}}$ is mapped to a softmax-equivalent logit pair (one
logit gauged to zero) and sampled per bit with the same top-$k$/top-$p$
truncation policy as our \textsc{Infinity}~\citep{han2025infinity}
implementation (the temperature is already applied in
Eq.~\eqref{eq:bitedit_raw}), recovering signs
$\tilde b^{k}_{p,j}=2u^{k}_{p,j}-1$ from labels $u^{k}_{p,j}\!\in\!\{0,1\}$. We
linearly anneal $\eta_{k}=\eta_{0}\bigl(1-(k-1)/(K-1)\bigr)$ from base scale
$\eta_{0}\!\geq\!0$, so guidance is strongest at the coarsest scale (global
layout and coarse semantics) and zero at the finest (high-frequency texture).
Setting $\eta_{0}\!=\!0$ recovers standard post-temperature CFG.

% --- 3.3 ResEdit -----------------------------------------------------------
\subsection{\ResEdit: Gated code-residual editing}
\label{sec:resedit}

\ResEdit{} acts on the combination layer: it takes as input the bits
sampled under \BitEdit{} (or any other guidance scheme) and determines
where and how strongly the implied code change is written into the
image's continuous latent, replacing discrete token reassembly with
per-scale \emph{code-space} residual injection.

\paragraph{Per-scale residual at the final latent resolution.}
Sampling under \BitEdit{} yields native-scale binary tokens
$\tilde b^{k} \!\in\! \{-1,+1\}^{D \times H_k \times W_k}$ and continuous
codes $\tilde c^{k} = \mathrm{BSQ}(\tilde b^{k})$. We upsample once and
form the residual at the final VAE resolution $H_z \!\times\! W_z$:
\begin{equation}
\label{eq:residual}
\bar c^{k} \!\triangleq\! \mathrm{up}_{k\to K}(\tilde c^{k}),
\quad
\bar c^{k}_{\mathrm{src}} \!\triangleq\! \mathrm{up}_{k\to K}(c^{k}_{\mathrm{src}}),
\quad
\Delta\bar c^{k} \!\triangleq\! \bar c^{k} - \bar c^{k}_{\mathrm{src}}
\;\in\; \mathbb{R}^{C \times H_z \times W_z},
\end{equation}
where $c^{k}_{\mathrm{src}}$ is the cached \emph{source-image} code from
the source pass (the source-pass BSQ encoding), rather than a same-prefix
source-conditional forward: the same-prefix forward is specific to
\BitEdit{} (Eq.~\eqref{eq:three_branch}), whereas the code-space residual
anchors preservation to the original image.

\paragraph{Localization mask.}
We assume a binary localization mask $M_{\mathrm{gt}}$ is provided
alongside the source--target prompt pair.\footnote{In benchmark experiments, we
use the dataset-provided ground-truth masks when available; otherwise, $M_{\mathrm{gt}}$ can be obtained
from an off-the-shelf grounding-and-segmentation pipeline.} The same
processed mask is shared across all scales: it is
dilated, bilinearly resized to $H_z \!\times\! W_z$ (typically a
downsample, since pixel-space masks are at the source-image resolution
while $H_z\!\times\!W_z$ is the tokenizer/VAE latent resolution),
Gaussian-blurred, and peak-normalized to $[0,1]$, yielding a soft-edged
spatial gate
\begin{equation}
\label{eq:gate_resolved}
G_{\mathrm{final}} \;\triangleq\; \mathrm{Process}(M_{\mathrm{gt}}) \;\in\; [0,1]^{H_z \times W_z}.
\end{equation}
This single soft mask lets us apply the same localization rule at all
scales, including coarse scales, without introducing a separate
hand-designed coarse-scale cutoff. Coarse scales can still influence
broad structure through the decoder and through the next-scale visual
prefix; the gate only constrains where the edited residual is written
into the cumulative latent.

\paragraph{The \ResEdit{} update.}
At the final resolution, the per-scale combination rule is
\begin{equation}
\label{eq:resedit}
\bar c^{k}_{\mathrm{out}} = \bar c^{k}_{\mathrm{src}} + \alpha\, G_{\mathrm{final}} \,\odot\, \Delta\bar c^{k},
\end{equation}
with $\odot$ a channel-broadcast spatial product and $\alpha\!\geq\!0$ a
residual-strength scalar. The edited image is
$\tilde x = \mathcal{D}\!\bigl(\sum_{k} \bar c^{k}_{\mathrm{out}}\bigr)$.
At background positions outside the processed edit region, where
$G_{\mathrm{final}}\!=\!0$, the update reduces to
$\bar c^{k}_{\mathrm{out}} = \bar c^{k}_{\mathrm{src}}$ for every scale
$k$, so the summed latent is exactly the source latent there.
Background \emph{latent features} are therefore preserved exactly; in
pixel space they are preserved up to boundary effects introduced by the
decoder receptive field and by the mask dilation/blur.

% --- 3.4 BitResEdit --------------------------------------------------------
\subsection{\BitResEdit: Coupling decision-time bits with combination-time codes}
\label{sec:bitresedit}

\BitEdit{} (log-odds, before sampling) and \ResEdit{} (code residuals,
after sampling) operate on different algebraic objects but are coupled
through the cumulative edited latent $\tilde z_{\le k}$, which is fed
back as the visual prefix for the next scale. The only other shared
state is the source-pass cache. Because each $\bar c^{k}_{\mathrm{out}}$
has been spatially gated at the final latent resolution, $\tilde z_{\le k}$
is an \emph{off-token-manifold} continuous prefix: it generally does not
correspond to any native-scale BSQ token sequence. We therefore feed
$\tilde z_{\le k}$ into the next scale through the same
downsampling/input-preparation operator \textsc{Infinity} uses to
construct the scale-$(k{+}1)$ visual input, treating \ResEdit{} as
intentional continuous code-space composition rather than discrete
token reassembly. Together \BitEdit{} and \ResEdit{} realize sampled
bit-level changes as localized, scale-aware code-space residuals
(Fig.~\ref{fig:method}).
\BitResEdit{}
requires no fine-tuning, no per-image optimization, and no auxiliary
generative editor; the localization mask is supplied by dataset
annotations or by an off-the-shelf grounding-and-segmentation model.
Algorithm~\ref{alg:edit} summarizes the edit pass in PyTorch-like
pseudocode: at each scale, the \BitEdit{} block (\S\ref{sec:bitedit})
guides and samples the bits, and the \ResEdit{} block
(\S\ref{sec:resedit}) writes the resulting code residual under the
mask; the Bernoulli-KL projection \texttt{kl\_clip} is detailed in
Appendix~\ref{sec:pseudocode} (Algorithm~\ref{alg:klclip}).

\begin{kalg}{\BitResEdit{}: edit pass}{\texttt{fwd} batches the three text conditions over one shared visual prefix; at $k{=}0$, each branch uses its own text-derived start map \texttt{sos}; \texttt{bsq}$(b)=b/\sqrt{D}$.}
\label{alg:edit}
\begin{lstlisting}[style=kaiming]
# fwd(p, c): per bit log odds head at prefix p, text c
# sample(d): per bit 2 class draw, Infinity top k/p truncation
# x_src: source image; S, T, null: source, target, uncond text
# M: edit mask; c_src = bsq_enc(x_src); G = process(M) in [0,1]
# Mb[k]: binary edit region gate from M at the scale k grid
# hw[k] = (Hk, Wk) scale grid; (Hz, Wz) = final latent grid
# hyperparams K, eta0, s, tau, delta, lim, alpha: see text

z = zeros(C, Hz, Wz)                 # cumulative edited latent
for k in range(K):
    eta = eta0 * (1 - k / (K - 1))   # anneal: strong coarse, 0 fine
    p = sos if k == 0 else to_input(z, hw[k])

    # BitEdit: decision time guidance on per bit log odds
    d_T, d_0, d_S = fwd(p, [T, null, S])     # 3 branch, shared prefix
    d_cfg = d_0 + s * (d_T - d_0)            # CFG
    a = d_cfg / tau                          # post temperature anchor
    d_raw = (d_cfg - eta * Mb[k] * (d_S - d_T)) / tau # source neg tilt
    d = kl_clip(d_raw, a, delta)             # Bernoulli KL ball
    d = a + clip(d - a, -lim, lim)           # secondary linf clamp
    b = 2 * sample(d) - 1                    # sample signed bits

    # ResEdit: combination time mask gated code residual
    c_e = up(bsq(b), (Hz, Wz))       # edited code, final res
    c_s = up(c_src[k], (Hz, Wz))     # source code, final res
    z = z + (c_s + alpha * G * (c_e - c_s))  # mask gated write

x_edit = decode(z)
\end{lstlisting}
\end{kalg}

%% file: chapters/exp.tex
\section{Experiments}

\subsection{Experimental setup}
\label{sec:setup}

\begin{table}[t]
\centering
\caption{PIE-Bench quantitative results. \BitResEdit{} attains the strongest text alignment among Infinity-2B editors while remaining competitive on background preservation. Methods are grouped by generative-model family: diffusion (top), flow matching (middle), and visual autoregressive (bottom). Best results are \textbf{bold}; second-best are \underline{underlined}.}
\label{tab:quantitative}
\setlength{\tabcolsep}{4pt}
\renewcommand{\arraystretch}{1.1}
\sisetup{table-format=2.2, mode=text, detect-weight=true, detect-family=true}
\resizebox{\textwidth}{!}{%
\begin{tabular}{l l S S[table-format=3.2] S[table-format=3.2] S S S}
\toprule
& & \multicolumn{4}{c}{Background Preservation} & \multicolumn{2}{c}{Text Alignment} \\
\cmidrule(lr){3-6} \cmidrule(lr){7-8}
{Method} & {Backbone} & {PSNR$\uparrow$} & {LPIPS$_{10^3}\downarrow$} & {MSE$_{10^4}\downarrow$} & {SSIM$_{10^2}\uparrow$} & {Whole$\uparrow$} & {Edited$\uparrow$} \\
\midrule
P2P~\citep{hertz2022prompt}        & SD1.4    & 17.87 & 208.80 & 219.88 & 71.14 & 25.01 & 22.44 \\
MasaCtrl~\citep{cao2023masactrl}   & SD1.4    & 22.17 & 106.62 &  86.97 & 79.67 & 23.96 & 21.16 \\
P2P-Zero~\citep{parmar2023zero}    & SD1.4    & 20.44 & 172.22 & 144.12 & 74.67 & 22.80 & 20.54 \\
NTI~\citep{mokady2023null}         & SD1.4    & 27.03 &  60.67 &  35.86 & 84.11 & 24.75 & 21.86 \\
PnP-Inv~\citep{ju2024pnp}          & SD1.4    & 22.46 & 106.06 &  80.45 & 79.68 & 25.41 & 22.62 \\
NP~\citep{Miyake_2025_WACV}        & SD1.4    & 26.21 &  69.01 &  39.73 & 83.40 & 24.61 & 21.87 \\
% SteerFlow
% Qwen-Image+RewardFlow
\midrule
StableFlow~\citep{Avrahami_2025_CVPR} & FLUX     & 21.64 &  92.28 & 115.21 & 84.94 & 24.65 & 21.70 \\
RF-Edit~\citep{wang2025taming}        & FLUX     & 23.22 & 131.18 &  75.00 & 81.44 & 25.22 & 22.40 \\
FlowEdit~\citep{Kulikov_2025_ICCV}    & FLUX     & 28.33 &  43.57 &  37.48 & 86.23 & 26.43 & 23.03 \\
VAGS~\citep{luo2026vags}              & SD3.5-L  & 26.38 &  70.38 &  34.86 & 87.68 & 26.92 & 23.08 \\
FlowChef~\citep{Patel_2025_ICCV}      & FLUX     & 29.03 &  43.11 &  36.67 & 87.44 & 27.05 & 23.09 \\
RewardFlow~\citep{susladkar2026rewardflow} & Qwen-Image-20B & \bfseries 32.09 & 38.47 & {\hphantom{0}\underline{23.57}} & 90.21 & \bfseries 29.78 & \bfseries 27.57 \\
\midrule
AREdit~\citep{wang2025aredit}     & Infinity-2B & 24.19 &  87.00 &  {--}  & 83.70 & 25.42 & 22.77 \\
MLN~\citep{elghoussani2026mln}    & SWITTI-2.5B   & {\underline{29.70}} & {\hphantom{0}\underline{36.50}} & {\hphantom{0}\bfseries 23.30} & 86.80 & 26.15 & 22.72 \\
VAREdit-8B~\citep{mao2025varedit}  & Infinity-8B & {--} & {--} & {--} & {--} & 26.60 & 23.30 \\
EditInfinity~\citep{wang2025editinfinity} & Infinity-2B & 27.95 & {\hphantom{0}\bfseries 33.08} & 24.27 & \bfseries 92.12 & 26.41 & 23.47 \\
\textbf{BitResEdit (ours)} & Infinity-2B & 26.67 & {\hphantom{0}43.21} & {\hphantom{0}31.61} & {\underline{91.71}} & {\underline{27.13}} & {\underline{24.54}} \\
\bottomrule
\end{tabular}%
}
\end{table}

\paragraph{Benchmark and metrics.}
We evaluate on PIE-Bench~\citep{ju2024pnp}, a standard text-guided image-editing
benchmark of $700$ source--target pairs covering $10$ editing categories
(random, change/add/delete object, attribute changes for content, pose, color
and material, change background, change style). Following the
benchmark protocol, we report two groups of metrics: (i) \emph{background
preservation} on the unedited region using PSNR, LPIPS~\citep{zhang2018perceptual},
MSE and SSIM~\citep{wang2004ssim}; and (ii) \emph{text alignment} using the
CLIP image--text similarity~\citep{radford2021clip} on the whole image and on
the edited region. All numbers are means over the full $700$-example set.

\paragraph{Backbone and configuration.}
\BitResEdit{} is training-free and runs on top of \textsc{Infinity-2B}~\citep{han2025infinity},
a bit-quantized residual VAR generator with $K\!=\!13$ scales and a $D$-bit
BSQ tokenizer. Across all experiments we keep the sampling temperature
$\tau$, top-$k$/top-$p$ truncation, CFG scale $s$, and KL bisection
budget $N\!=\!4$ at fixed defaults (values in
Appendix~\ref{app:impl-details}), and report sensitivity to the main
\BitEdit{} and \ResEdit{} components and to mask source in
\S\ref{sec:ablation}. For Table~\ref{tab:quantitative} we use, as
the localization mask, the binary \emph{ground-truth mask} provided with
each PIE-Bench source--target pair, which gives the strongest
background-preservation profile among the mask sources we test
(Table~\ref{tab:ablation:mask}). Robustness to weaker mask sources,
including a tight axis-aligned bounding box of the same region and an
automatic GroundingDINO+SAM~\citep{liu2024groundingdino,kirillov2023sam}
pipeline, is reported in \S\ref{sec:ablation:mask}.

\paragraph{Baselines.}
Figure~\ref{fig:per-category-radar} compares \BitResEdit{} against
three VAR editors and two flow-matching editors.
AREdit~\citep{wang2025aredit} edits training-free on Infinity-2B by
recombining cached source and target token streams under the mask
with attention refinement. VAREdit-8B~\citep{mao2025varedit} trains a
next-scale editing model with scale-aligned source conditioning on
Infinity-8B. EditInfinity~\citep{wang2025editinfinity} inverts each
source image by per-image fine-tuning of Infinity-2B before editing.
FlowChef~\citep{Patel_2025_ICCV} regenerates the image with FLUX,
steering each flow-matching step toward the target prompt with
gradient updates. VAGS~\citep{luo2026vags} extends
FlowEdit~\citep{Kulikov_2025_ICCV} on SD3.5-L with a
velocity-adaptive guidance scale and takes no localization mask.
Except for Table~\ref{tab:quantitative}, which quotes numbers from
the original papers, all tables and figures report our own runs of
the released models or our re-implementations
(Appendix~\ref{app:impl-details}).

\subsection{Quantitative results}
\label{sec:quant}

\paragraph{PIE-Bench leaderboard.}
Table~\ref{tab:quantitative}\footnote{All numbers in
Table~\ref{tab:quantitative} are taken from the original papers' reported
results or from RewardFlow~\citep{susladkar2026rewardflow}. All other
tables and figures report our own runs of the released models or our
re-implementations; running details and the full-set metrics of these
runs are in Appendix~\ref{app:impl-details}
(Table~\ref{tab:baseline-repro}).} summarizes background-preservation and
text-alignment metrics across all three families. Under the default
GT-mask localization, \BitResEdit{} is second-best overall on SSIM and
on both CLIP scores, behind only the same-backbone EditInfinity on SSIM
and the $20$B-parameter RewardFlow on text alignment. Within the
Infinity-2B backbone, \BitResEdit{} attains the strongest text
alignment, improving CLIP-Edited by $+1.07$ over the best prior
same-backbone editor (EditInfinity) while staying within $0.42$ SSIM of
its background preservation. PSNR is \BitResEdit{}'s weakest metric: it
trails MLN, EditInfinity, and the strongest flow-matching and inversion
baselines, and among Infinity-2B editors it improves only over AREdit.

\begin{figure}[t]
  \centering
  \begin{minipage}[t]{0.32\linewidth}
    \centering
    \includegraphics[width=\linewidth]{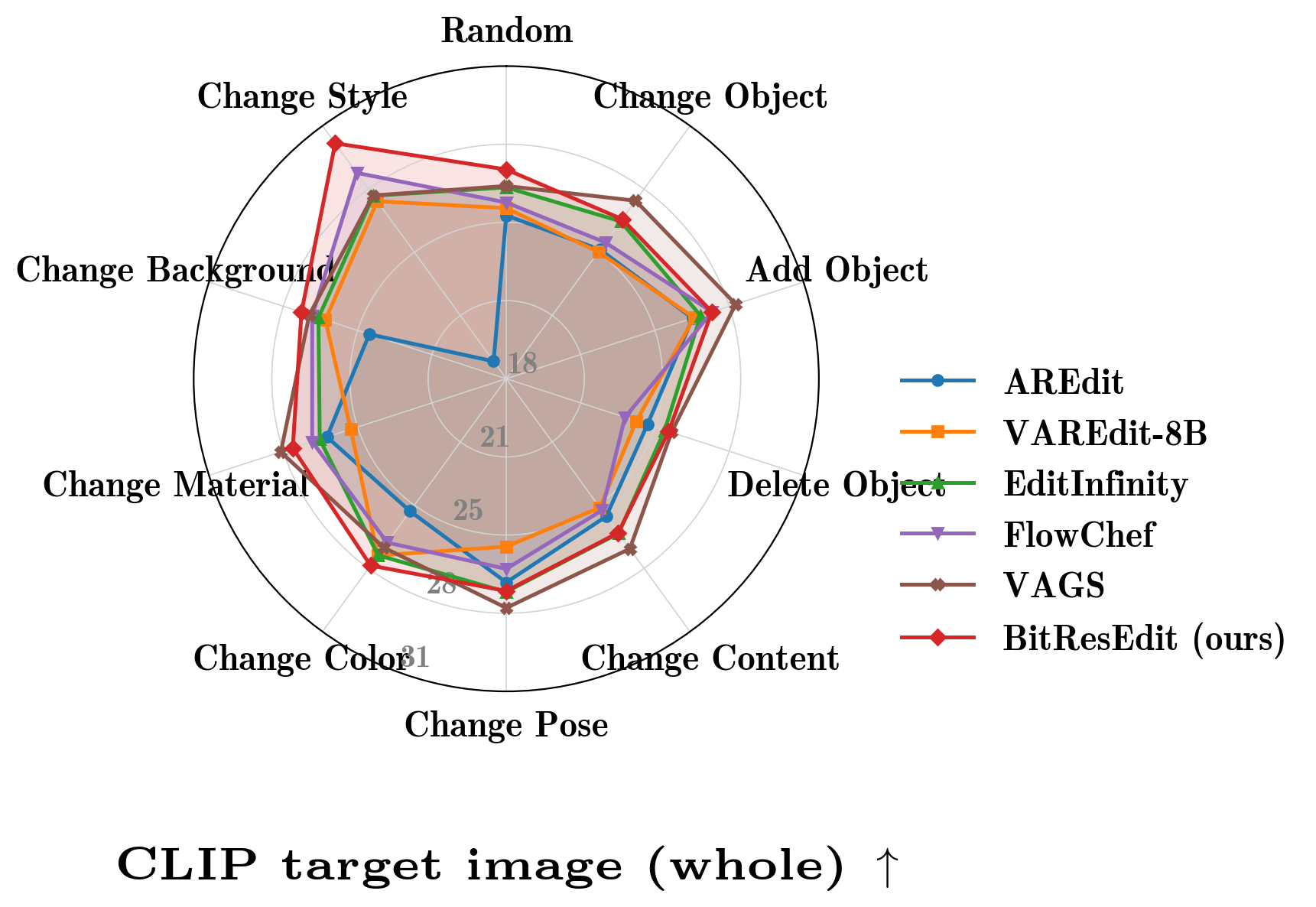}
  \end{minipage}\hfill
  \begin{minipage}[t]{0.32\linewidth}
    \centering
    \includegraphics[width=\linewidth]{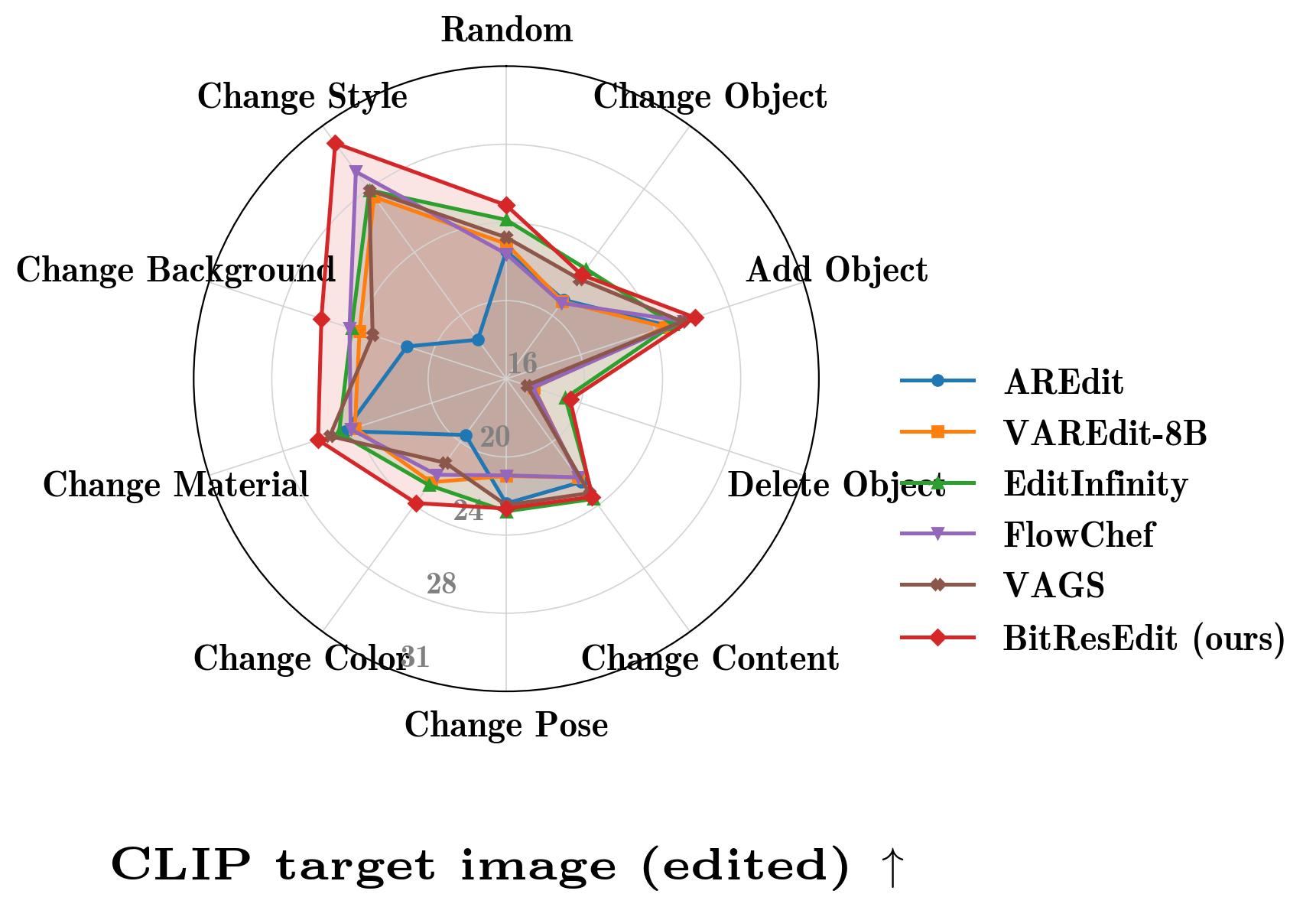}
  \end{minipage}\hfill
  \begin{minipage}[t]{0.32\linewidth}
    \centering
    \includegraphics[width=\linewidth]{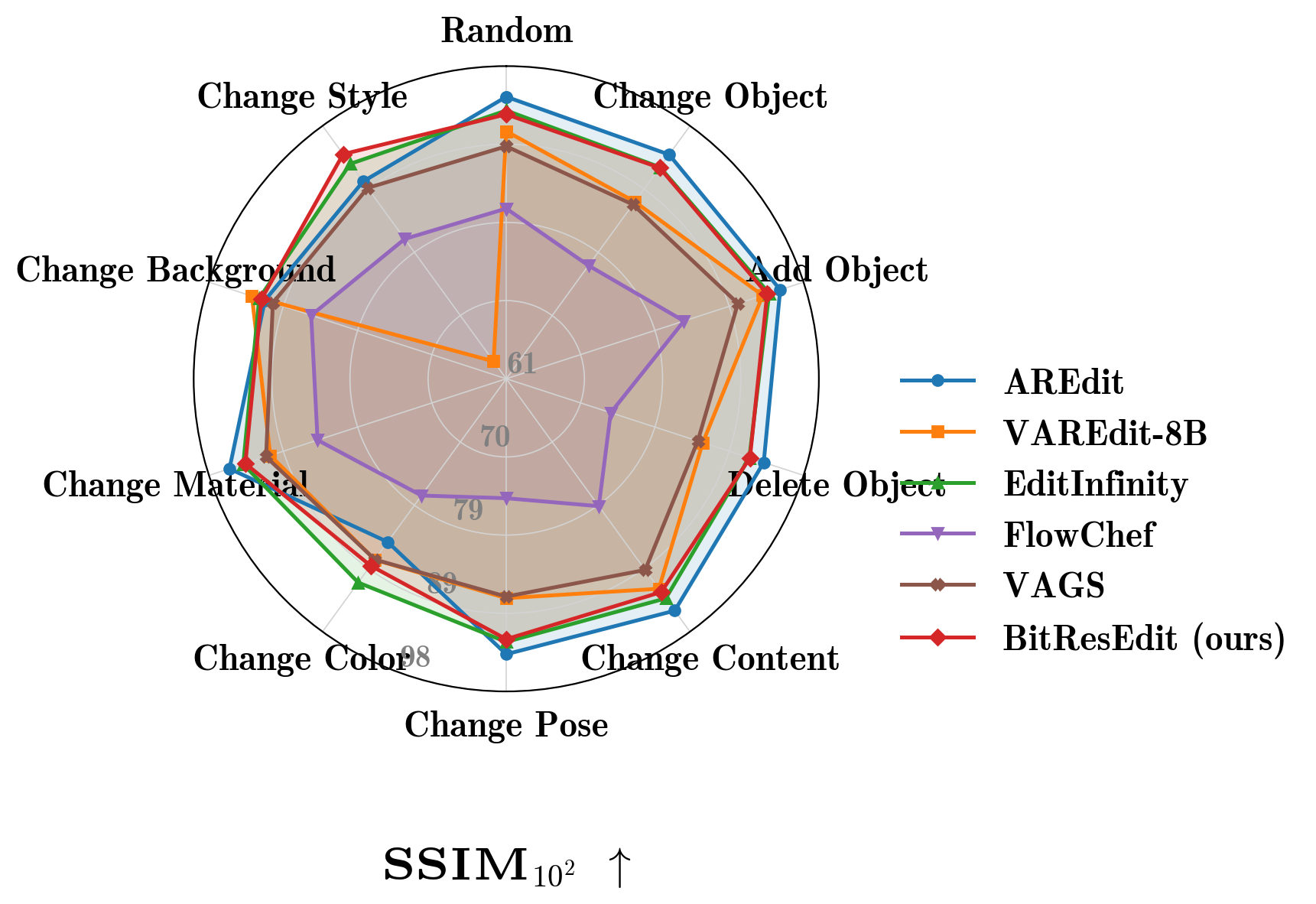}
  \end{minipage}
  \caption{Per-category PIE-Bench results for AREdit, VAREdit-8B, EditInfinity, FlowChef and VAGS (our reproductions; Appendix~\ref{app:impl-details}), and \BitResEdit{}. \BitResEdit{} encloses the largest CLIP polygons while its SSIM polygon stays compact across all ten categories. All methods use GT-mask localization except VAREdit-8B, whose edits are driven by the instruction alone, and the mask-free VAGS, whose edits are driven by the prompt pair alone.}
  \label{fig:per-category-radar}
\end{figure}

% Component-ablation table for \S\ref{sec:ablation:components}; placed here so
% it lands between the per-category radar and qualitative figures.
\providecommand{\cmark}{\ensuremath{\checkmark}}
\providecommand{\xmark}{\ensuremath{\text{\sffamily x}}}
\begin{table}[t]
\centering
\caption{Component ablation on PIE-Bench. \ResEdit{} carries background
preservation, \BitEdit{} carries text alignment, and their coupling is
second-best on all six metrics. \BitEdit{} and \ResEdit{} are toggled
with other settings fixed.}
\label{tab:ablation:components}
\setlength{\tabcolsep}{4pt}
\renewcommand{\arraystretch}{1.1}
\sisetup{table-format=2.2, mode=text, detect-weight=true, detect-family=true}
\resizebox{\textwidth}{!}{%
\begin{tabular}{l c c S S[table-format=3.2] S[table-format=3.2] S S S}
\toprule
& & & \multicolumn{4}{c}{Background Preservation} & \multicolumn{2}{c}{Text Alignment} \\
\cmidrule(lr){4-7} \cmidrule(lr){8-9}
{Configuration} & {\BitEdit{}} & {\ResEdit{}}
  & {PSNR$\uparrow$} & {LPIPS$_{10^3}\downarrow$} & {MSE$_{10^4}\downarrow$}
  & {SSIM$_{10^2}\uparrow$} & {Whole$\uparrow$} & {Edited$\uparrow$} \\
\midrule
baseline  (w/ GT)                                  & \xmark & \xmark & 24.21 & 68.90 & 70.40 & 88.58 & 23.01 & 19.32 \\
\quad +\,\BitEdit{} only                            & \cmark & \xmark & 11.66 & 402.15 & 908.46 & 55.17 & 27.89 & 24.66 \\
\quad +\,\ResEdit{} only                            & \xmark & \cmark & 29.46 & 31.83 & 17.40 & 93.10 & 26.11 & 23.17 \\
\BitResEdit{} (ours)                                & \cmark & \cmark & 26.67 & 43.21 & 31.61 & 91.71 & 27.13 & 24.54 \\
\bottomrule
\end{tabular}%
}
\end{table}

\paragraph{Preservation, alignment, and latency.}
The leaderboard exposes a clear trade-off: inversion (NTI, NP) and flow
(FlowEdit, FlowChef) baselines sit at moderate preservation and
alignment, while RewardFlow buys top alignment at a $20$B-parameter
cost. Within Infinity-2B, EditInfinity leads preservation and
\BitResEdit{} leads alignment, reflecting our two-stage design:
\ResEdit{}'s residual injection clamps unedited regions to the source,
while \BitEdit{}'s Bernoulli-KL guidance drives the CLIP gains.
The trade-off extends to latency (Table~\ref{tab:latency},
Appendix~\ref{app:impl-details}): EditInfinity's preservation lead
costs per-image fine-tuning, which makes it $\sim\!45\times$ slower
than \BitResEdit{}, and RewardFlow's alignment lead costs
reward-guided sampling on its $20$B backbone. \BitResEdit{} edits one
image in $10.41$\,s, in line with the training-free VAR baselines.

\paragraph{Per-category behavior.}
Figure~\ref{fig:per-category-radar} breaks the comparison down over
PIE-Bench's $10$ categories. On CLIP alignment, \BitResEdit{}
encloses the largest polygon on both panels: it surpasses AREdit and
VAREdit-8B in every category, and its few losses to the
same-backbone EditInfinity are within $0.36$. FlowChef trails in
nearly every category and its SSIM polygon is the innermost: it
regenerates the full image from noise with soft guidance rather than
anchoring unedited regions. The mask-free VAGS is the stronger flow
baseline, and even edges \BitResEdit{} on CLIP-Whole in six of ten
categories. But it trails on CLIP-Edited and SSIM$_{10^2}$ in all
ten: without localization, its edits bleed into unedited regions. On
SSIM$_{10^2}$, AREdit preserves best on the seven local-edit
categories but degrades sharply on the three global ones, where
\BitResEdit{} leads; EditInfinity edges \BitResEdit{} on most
categories, but by under $2.5$ points. The hardest category is
\emph{change style}: VAREdit-8B's SSIM collapses to $63.58$ while
\BitResEdit{} holds at $93.67$. The polygons are compact and
centered: the gains are not driven by a few easy categories.

\begin{figure}[!ht]
  \centering
  \includegraphics[width=\linewidth]{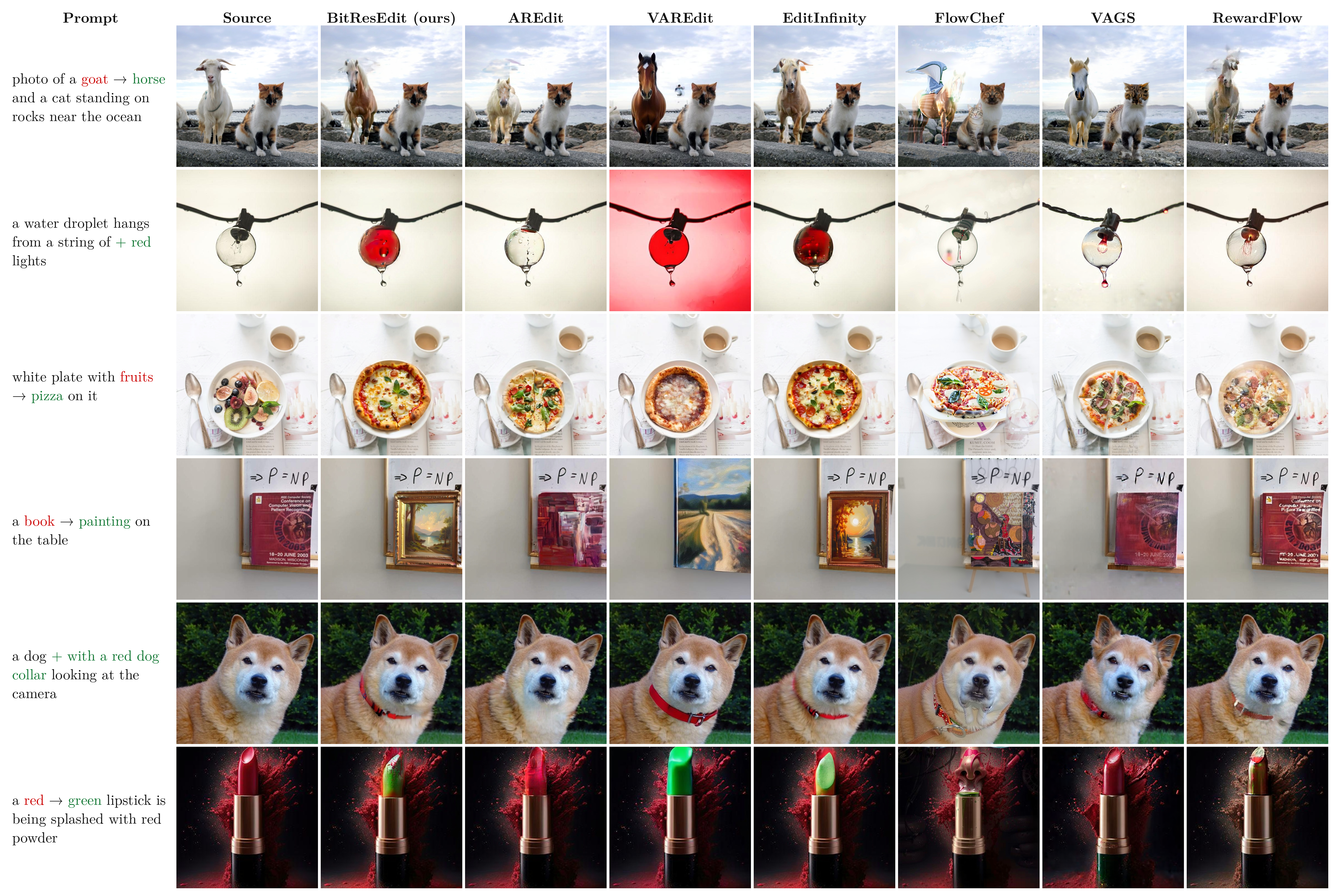}
  \caption{Qualitative comparison on PIE-Bench. Columns show the source image and edits produced by \BitResEdit{}, AREdit, VAREdit, EditInfinity, FlowChef, VAGS, and RewardFlow for the prompt on the left of each row. \BitResEdit{} applies the requested change while anchoring unedited regions to the source.}
  \label{fig:qualitative-comparison}
\end{figure}

\subsection{Qualitative results}

Figure~\ref{fig:qualitative-comparison} compares \BitResEdit{} against AREdit, VAREdit, EditInfinity, FlowChef, VAGS, and RewardFlow on a representative set of PIE-Bench prompts spanning object swap, attribute change, and color edits. \BitResEdit{} applies the requested change while preserving non-target regions (e.g.\ background, layout, and surrounding objects), whereas the baselines either drift in identity, leak edits into untargeted areas, or under-apply the prompt. These failure modes track the quantitative trade-off. In the goat-to-horse swap, \BitResEdit{} produces a natural horse and leaves the cat, the rocks, and the ocean untouched. VAGS produces a comparable horse but changes the cat as well. Without a mask, its edits bleed into unedited regions in every row; in the collar edit it replaces the dog outright. FlowChef leaves distorted artifacts in the horse swap and several other rows. When the string lights turn red, the instruction-driven VAREdit lets the color flood the whole frame and FlowChef drifts in global tone. When the lipstick turns green, most of the methods leave it red, and EditInfinity's and VAREdit's flat green clashes with the red powder around it. \BitResEdit{} applies both color edits inside the mask and nowhere else, the behavior \ResEdit{}'s gated residual enforces by construction.

\subsection{Ablation study}
\label{sec:ablation}
\input{chapters/ablation.tex}

%% file: chapters/ablation.tex
% Ablation tables for Section~\ref{sec:ablation}. Numeric cells marked \tbd{}
% are placeholders to be filled from PIE-Bench v1 runs (one row per editor
% configuration; seed, CFG temperature, top-k/p, and mask processing held
% constant across rows so deltas are attributable to the toggled component).
% Column layout matches Table~\ref{tab:quantitative} in chapters/exp.tex.
\providecommand{\tbd}{{\color{gray}--}}
\providecommand{\cmark}{\ensuremath{\checkmark}}
\providecommand{\xmark}{\ensuremath{\text{\sffamily x}}}

\subsubsection{Independent \BitEdit{} \& \ResEdit{}}
\label{sec:ablation:components}

% Table~\ref{tab:ablation:components} lives in chapters/exp.tex (between the
% per-category radar and qualitative figures) to control float placement.

We disentangle the two contributions of \BitResEdit{} with a $2{\times}2$
factorial on PIE-Bench. Each row of Table~\ref{tab:ablation:components}
toggles \BitEdit{} (log-odds-level
source-negative guidance with Bernoulli-KL projection,
Eqs.~\eqref{eq:bitedit_raw}--\eqref{eq:bitedit_alpha}) and \ResEdit{} (gated
code-space residual injection at the final latent resolution,
Eq.~\eqref{eq:resedit}) independently. ``No \BitEdit{}'' rows fall back to
standard post-temperature CFG (the 3-branch forward collapses to 2-branch);
``no \ResEdit{}'' rows accumulate the \BitEdit{}-sampled target codes directly
into the latent without per-scale gated residual or mask compositing, so
source enters only through \BitEdit{}'s logit-space source-negative direction.
The numbers reveal a sharp division of labor: \ResEdit{} owns every
background-preservation metric (PSNR/LPIPS/MSE/SSIM), \BitEdit{} owns every
text-alignment metric (CLIP-Whole/Edited), and \BitResEdit{} is second-best on
all six --- close to each component's strength on its own axis without
sacrificing the other.

\begin{table}[t]
\centering
\caption{\BitEdit{} sub-knob ablation on PIE-Bench, with \ResEdit{} enabled.
The $\eta_{k}$ annealing matters most; the KL trust region rarely binds. Each row
turns off exactly one \BitEdit{} mechanism. The KL trust-region row uses
$\delta_{\mathrm{KL}}{=}2.0$ (the largest value we swept; the projection is
essentially inactive at this radius and serves as an upper bound for the
$\delta_{\mathrm{KL}}{\to}\infty$ limit).}
\label{tab:ablation:bitedit-inner}
\setlength{\tabcolsep}{4pt}
\renewcommand{\arraystretch}{1.1}
\sisetup{table-format=2.2, mode=text, detect-weight=true, detect-family=true}
\resizebox{\textwidth}{!}{%
\begin{tabular}{l S S[table-format=3.2] S[table-format=3.2] S S S}
\toprule
& \multicolumn{4}{c}{Background Preservation} & \multicolumn{2}{c}{Text Alignment} \\
\cmidrule(lr){2-5} \cmidrule(lr){6-7}
{\BitEdit{} variant}
  & {PSNR$\uparrow$} & {LPIPS$_{10^3}\downarrow$} & {MSE$_{10^4}\downarrow$}
  & {SSIM$_{10^2}\uparrow$} & {Whole$\uparrow$} & {Edited$\uparrow$} \\
\midrule
\textbf{Full \BitEdit{}} ($\eta_{0}{>}0$, KL proj., source-neg., annealed)
                                              & 26.67 & 43.10 & 31.59 & 91.73 & \bfseries 27.15 & \bfseries 24.58 \\
\;\;-- trust region ($\delta_{\mathrm{KL}}{=}2.0$)
                                              & 26.68 & 43.11 & 31.51 & 91.72 & 27.08 & 24.55 \\
\;\;-- annealing ($\eta_{k}{=}\eta_{0}$ constant)
                                              & \bfseries 27.04 & 43.52 & \bfseries 28.74 & 91.51 & 26.62 & 23.37 \\
\;\;-- source-negative ($\Delta^{k}$ EP anchor)
                                              & 26.77 & 42.64 & 30.65 & \bfseries 91.80 & 26.90 & 24.42 \\
\;\;-- guidance ($\eta_{0}{=}0$; $\equiv$ standard CFG)
                                              & 26.79 & \bfseries 42.59 & 30.50 & \bfseries 91.80 & 26.89 & 24.43 \\
\;\;-- \BitEdit{} entirely (\texttt{arcfg\_mode}=off)
                                              & 25.90 & 45.45 & 38.43 & 91.37 & 26.38 & 23.51 \\
\bottomrule
\end{tabular}%
}
\end{table}

\begin{table}[t]
\centering
\caption{\ResEdit{} sub-knob ablation on PIE-Bench, with \BitEdit{} enabled.
The localization gate $G_{\mathrm{final}}$ holds the preservation--alignment
trade-off in place. The $\alpha{=}1.3$ row is the closest tested value to the
symbolic $\alpha{=}1.5$ in the original method specification.}
\label{tab:ablation:resedit-inner}
\setlength{\tabcolsep}{4pt}
\renewcommand{\arraystretch}{1.1}
\sisetup{table-format=2.2, mode=text, detect-weight=true, detect-family=true}
\resizebox{\textwidth}{!}{%
\begin{tabular}{l S S[table-format=3.2] S[table-format=3.2] S S S}
\toprule
& \multicolumn{4}{c}{Background Preservation} & \multicolumn{2}{c}{Text Alignment} \\
\cmidrule(lr){2-5} \cmidrule(lr){6-7}
{\ResEdit{} variant}
  & {PSNR$\uparrow$} & {LPIPS$_{10^3}\downarrow$} & {MSE$_{10^4}\downarrow$}
  & {SSIM$_{10^2}\uparrow$} & {Whole$\uparrow$} & {Edited$\uparrow$} \\
\midrule
\textbf{Full \ResEdit{}} ($\alpha{=}1$, gated, dilated mask)
                                              & 26.67 & 43.10 & 31.59 & 91.73 & \bfseries 27.15 & \bfseries 24.58 \\
\;\;-- localization ($G_{\mathrm{final}}{\equiv}1$)
                                              & 29.49 & 40.32 & 27.40 & 91.04 & 24.45 & 21.21 \\
\;\;-- mask dilation (kernel${=}0$)
                                              & 29.15 & 33.75 & 18.52 & 93.22 & 26.95 & 24.44 \\
\;\;$\alpha{=}0.5$                            & 31.09 & \bfseries 27.90 & 11.47 & \bfseries 93.97 & 25.68 & 22.87 \\
\;\;$\alpha{=}1.3$ (proxy for $\alpha{=}1.5$) & 25.18 & 50.57 & 44.33 & 90.56 & 27.01 & 24.18 \\
\bottomrule
\end{tabular}%
}
\end{table}

\paragraph{Inner ablation of \BitEdit{}.}
Within \BitEdit{}, we further isolate the source-negative direction
$\Delta^{k}$ (Eq.~\eqref{eq:neg_direction}), the Bernoulli-KL trust-region
projection (Eq.~\eqref{eq:bitedit_alpha}), and the linear $\eta_{k}$ schedule
(\S\ref{sec:bitedit}). Each row in Table~\ref{tab:ablation:bitedit-inner}
turns off exactly one mechanism while keeping \ResEdit{} fully enabled, so the
remaining edit pressure flows through the gated code-space residual.\footnote{The
\textbf{Full} anchor row shared by Tables~\ref{tab:ablation:bitedit-inner}
and~\ref{tab:ablation:resedit-inner} (CLIP-Whole $27.15$) is the reference run of
this ablation campaign, against which the within-table deltas are measured. It is a
\emph{separate} run from the identically-configured production model reported in
Table~\ref{tab:quantitative} (CLIP-Whole $27.13$): the two use the same configuration
and seed, and the $0.02$ gap reflects multi-GPU sampling nondeterminism, well within
the $\pm0.04$~CLIP-T seed-noise floor estimated for this benchmark.}
Two findings stand out. (i) The KL trust-region projection is essentially
inactive at our default $\delta_{\mathrm{KL}}{=}1.0$: widening to
$\delta_{\mathrm{KL}}{=}2.0$ shifts CLIP-T by $-0.07$ and PSNR by $+0.01$,
both within ${\approx}2\sigma$ of the seed-noise floor ($\pm0.04$~CLIP-T,
$\pm0.02$~PSNR; estimated from three baseline seeds), so the bisection
rarely binds on the production overlay and only acts as a safety rail for
occasional outlier bits. (ii) The other four ablations each cost between
$0.25$ and $0.77$~CLIP-Whole when disabled. Removing
$\eta_{k}$-annealing collapses CLIP-Edited the most ($-1.21$), confirming
that concentrating guidance pressure at the coarsest scales is what drives
prompt-aligned semantic changes. Dropping the source-negative anchor (the
``EP'' variant, in which $\Delta^{k}$ is taken against $\varnothing$ rather
than the post-CFG target) and zeroing $\eta_{0}$ collapse to almost
identical metrics, because both leave the per-bit log-odds at the
post-temperature CFG anchor; this empirically validates the Eq.~\eqref{eq:bitedit_raw}
prediction that $\eta_{0}{=}0$ recovers standard CFG. Disabling
\BitEdit{} altogether (\texttt{arcfg\_mode=off}) is the worst row on every
preservation metric as well, since the source-conditional branch is no longer
used to keep edits localized in log-odds space.

\begin{table}[t]
\centering
\caption{Mask-source robustness on PIE-Bench. An automatic
GroundingDINO+SAM mask tracks the GT mask closely; a bounding box
collapses preservation. All other settings use the default
\BitResEdit{} configuration.}
\label{tab:ablation:mask}
\setlength{\tabcolsep}{4pt}
\renewcommand{\arraystretch}{1.1}
\sisetup{table-format=2.2, mode=text, detect-weight=true, detect-family=true}
\resizebox{\textwidth}{!}{%
\begin{tabular}{l S S[table-format=3.2] S[table-format=3.2] S S S}
\toprule
& \multicolumn{4}{c}{Background Preservation} & \multicolumn{2}{c}{Text Alignment} \\
\cmidrule(lr){2-5} \cmidrule(lr){6-7}
{Mask source}
  & {PSNR$\uparrow$} & {LPIPS$_{10^3}\downarrow$} & {MSE$_{10^4}\downarrow$}
  & {SSIM$_{10^2}\uparrow$} & {Whole$\uparrow$} & {Edited$\uparrow$} \\
\midrule
GT mask (default)                          & \bfseries 26.67 & \bfseries 43.21 & \bfseries 31.61 & \bfseries 91.71 & 27.13 & 24.54 \\
Bounding box of GT region                  & 19.68 & 130.07 & 233.10 & 82.27 & \bfseries 27.35 & \bfseries 24.70 \\
GroundingDINO + SAM (auto)                 & 25.75 & 47.82 & 42.59 & 90.77 & 26.52 & 23.65 \\
% No mask ($G_{\mathrm{final}}{\equiv}1$)    & {\tbd} & {\tbd} & {\tbd} & {\tbd} & {\tbd} & {\tbd} \\
\bottomrule
\end{tabular}%
}
\end{table}

\paragraph{Inner ablation of \ResEdit{}.}
Within \ResEdit{}, we isolate the localization gate $G_{\mathrm{final}}$
(Eq.~\eqref{eq:gate_resolved}), the mask dilation that softens its
boundary (\S\ref{sec:resedit}), and the residual-strength scalar $\alpha$
(Eq.~\eqref{eq:resedit}). Each row in Table~\ref{tab:ablation:resedit-inner}
keeps \BitEdit{} fully enabled and toggles one ingredient of \ResEdit{}.
The clearest signal is that $G_{\mathrm{final}}$ is what holds the
preservation--alignment trade-off in place: removing the mask
($G_{\mathrm{final}}{\equiv}1$) drops both CLIP-Whole ($-2.69$) and
CLIP-Edited ($-3.37$) by large margins because the edited residual now
writes everywhere, including outside the prompt's target region; even
though pixel-level PSNR goes up, the resulting image disagrees with the
text. Skipping the mask-dilation step (kernel~$=0$) keeps the mask alive
but lets the boundary stay tight to the GT annotation: this preserves
CLIP-Whole within $0.20$ of the baseline while picking up an additional
$\sim\!2.5$~PSNR and $\sim\!9$~LPIPS$_{10^3}$, indicating that the
dilation buffer in our processed gate exists primarily to soak up small
GT-mask annotation errors rather than to improve preservation. The
$\alpha$ sweep behaves as a smooth knob: $\alpha{=}0.5$ defines the
preservation-extreme of our front (PSNR~$=31.09$, SSIM~$=93.97$) at the
cost of $1.47$~CLIP-Whole; $\alpha{=}1.3$ (our closest tested value to
$\alpha{=}1.5$) over-edits, dropping PSNR by $1.49$ for a marginal
CLIP-Whole change.

\subsubsection{Robustness to mask}
\label{sec:ablation:mask}

\BitResEdit{} consumes a binary localization mask $M_{\mathrm{gt}}$ to
construct $G_{\mathrm{final}}$. We test sensitivity to the mask source by
swapping the dataset's ground-truth mask with two weaker substitutes: a
mask predicted by an off-the-shelf grounding-and-segmentation pipeline
(GroundingDINO + SAM), and a tight axis-aligned bounding box around the
same region. Table~\ref{tab:ablation:mask} shows that the automatic
GroundingDINO+SAM pipeline tracks the GT-mask configuration closely,
losing only $0.92$ PSNR, $0.94$ SSIM, and $0.89$ CLIP-Edited points,
which indicates that \BitResEdit{} does not require an oracle mask to
retain most of its preservation--alignment profile. The tight bounding
box, in contrast, slightly improves CLIP-Whole ($+0.22$) and CLIP-Edited
($+0.16$) but collapses preservation by $\sim\!7$ PSNR, $\sim\!87$ LPIPS
and $\sim\!9$ SSIM, since the rectangular envelope leaks edits into the
unmasked corners around the target region. Mask \emph{tightness}, not
perfect boundaries, therefore drives the preservation--alignment
trade-off in our pipeline.

%% file: chapters/related_work.tex
\section{Related Work}
\subsection{Visual autoregressive models}
Visual autoregressive (AR) models generate images by factorizing visual data
into ordered predictions, from raw pixels to discrete or continuous latent
variables. Early work such as iGPT~\citep{chen2020generative},
VQ-VAE~\citep{oord2017neural}, and VQGAN~\citep{esser2021taming} established
next-token prediction over pixels or quantized latents, while
DALL-E~\citep{ramesh2021zero} and Parti~\citep{yu2022scaling} scaled this
paradigm to text-conditioned image generation. Recent visual AR models revisit
the ordering and representation of visual tokens. VAR~\citep{tian2024visual}
replaces raster-scan prediction with coarse-to-fine next-scale prediction, and
LlamaGen~\citep{sun2024autoregressive} shows that standard LLaMA-style AR
transformers remain competitive when paired with strong visual tokenizers.
Infinity~\citep{han2025infinity} further scales next-scale AR through bitwise
token prediction for high-resolution synthesis. Follow-up works improve
flexibility, efficiency, or representation: FlowAR~\citep{ren2024flowar}
combines scale-wise AR with flow matching; M-VAR~\citep{ren2024mvar} decouples
intra- and inter-scale modeling; FlexVAR~\citep{jiao2025flexvar} removes
residual prediction; HMAR~\citep{kumbong2025hmar} adds masked generation;
Continuous VAR~\citep{shao2025continuous} avoids vector quantization; and
InfinityStar~\citep{liu2025infinitystar} extends discrete AR modeling to
unified image-video generation. Our work builds on this line: we keep a
bitwise-residual generator, \textsc{Infinity}, frozen and use two structures
native to it---the per-bit prediction head and the additive multi-scale code
field---as the interface for editing.

\subsection{Image editing for visual autoregressive models}
% AREdit~\citep{wang2025training} presents a training-free Visual Autoregressive (VAR) framework that
% achieves high-fidelity local edits by caching the original image's token
% distributions and utilizing adaptive fine-grained masking alongside a token
% re-assembling strategy.
% VAREdit~\citep{varedit2026} proposes a tuning-based approach that
% formulates instruction-guided image editing as a multi-scale conditional
% generation task. To overcome scale-mismatch issues between source details and
% target features, it introduces a Scale-Aligned Reference (SAR) module that
% injects scale-matched source conditions specifically into the model's first
% self-attention layer. EditInfinity~\citep{wangeditinfinity} adapts binary-quantized generative models
% via a parameter-efficient approach that leverages exact intermediate quantized
% representations as supervision for precise image inversion. The method
% optimizes a learnable prompt and LoRA for style preservation, employing a
% piecewise linear smoothing kernel to seamlessly blend source and target tokens
% during multi-scale autoregressive editing. Masked Logit Nudging (MLN;~\citealp{el2026prompt}) offers
% an architecture-agnostic, inversion-free technique that operates directly in
% logit space by softly nudging the model's predictions toward the source image's
% token distribution. MLN restricts modifications using spatial masks derived
% from cross-attention differences and introduces a quantization refinement step
% to restore background details and correct discretization artifacts.

Diffusion-based image editing has achieved strong results through denoising
priors, inversion, and attention control, including
SDEdit~\citep{meng2021sdedit}, Prompt-to-Prompt~\citep{hertz2022prompt},
Null-text Inversion~\citep{mokady2023null}, DiffEdit~\citep{couairon2022diffedit},
Plug-and-Play Diffusion Features~\citep{tumanyan2023plug},
Pix2Pix-Zero~\citep{parmar2023zero}, MasaCtrl~\citep{cao2023masactrl}, and
InstructPix2Pix~\citep{brooks2023instructpix2pix}. Flow-matching backbones
extend this line with inversion-free editors:
FlowEdit~\citep{Kulikov_2025_ICCV} transports the source latent directly
toward the target prompt, and VAGS~\citep{luo2026vags} augments it with a
velocity-adaptive guidance scale that modulates the CFG strength per step
from the alignment between source and target velocity fields. All of these
methods, however, rely on iterative denoising rather than causal
visual-token generation. \BitResEdit{} shares the inversion-free, guidance-based
spirit of this line, but realizes it in the discrete bitwise sampler of a VAR
generator, where there is no continuous trajectory to steer.

In contrast, image editing for visual autoregressive models remains
under-explored, but offers a token-level alternative that can naturally exploit
causal generation, discrete visual codes, and coarse-to-fine prediction.
EditAR~\citep{mu2025editar} formulates editing and other conditional generation
tasks as unified next-token prediction over visual tokens.
AREdit~\citep{wang2025aredit} performs training-free text-guided editing with
VAR by caching source token indices and probability distributions, while
VAREdit~\citep{mao2025varedit} trains a next-scale editing model with
scale-aligned source conditioning. Recent works further improve controllability
and post-training: EARL~\citep{ahmadi2025earl} applies reinforcement learning
to autoregressive image editing, Masked Logit
Nudging~\citep{elghoussani2026mln} preserves unchanged regions through masked
source-token guidance, and Rethinking Structure
Preservation~\citep{xia2026structure} injects structure-related VAR features
for better layout consistency. VARIN~\citep{dao2025varin} inverts the discrete
Gumbel noise of next-scale sampling through a location-aware argmax
pseudo-inverse, reconstructing the source image and applying prompt-guided edits
while preserving background structure. These editors intervene on token
streams, learned conditioning, token-level logits, features, or sampling noise.
\BitResEdit{} instead intervenes at two points these methods leave untouched:
\BitEdit{} acts before sampling, on the per-bit Bernoulli log-odds under a
closed-form KL trust region, and \ResEdit{} acts after sampling, composing
mask-gated residuals in the continuous code field from which the image is
assembled---all training-free, with a frozen generator.

%% file: chapters/conclusion.tex
\section{Conclusion}
\label{sec:conclusion}

We presented \BitResEdit{}, a training-free editor that exploits the native
geometry of bitwise-residual VAR generators: \BitEdit{} performs bounded
source-negative guidance in per-bit Bernoulli log-odds space, while \ResEdit{}
writes the resulting changes as mask-gated residuals in the continuous
sum-of-scales code space. On PIE-Bench, this coupling gives strong target
alignment among same-backbone Infinity-2B editors while retaining competitive
background preservation, and ablations show that the two components play
complementary roles. \BitResEdit{} touches no weights and adds no auxiliary
model; the editor is the generator's own bit distributions and code
arithmetic. We hope this perspective---edit the bits, diff the codes---carries
over to other bit-quantized autoregressive generators.

%% file: chapters/implementation_details.tex
% ---------------------------------------------------------------------------
% Appendix: running details for BitResEdit and for the baselines we
% measure ourselves (AREdit, VAREdit-8B, EditInfinity, FlowChef, VAGS,
% RewardFlow), plus the full-set aggregates of those runs
% (tab:baseline-repro) and the inference-latency analysis and table
% (tab:latency). Referenced from the quantitative table
% (tab:quantitative), the per-category
% radar (fig:per-category-radar), and the qualitative comparison
% (fig:qualitative-comparison).
% ---------------------------------------------------------------------------
\section{Implementation Details}
\label{app:impl-details}

\paragraph{\BitResEdit{}.}
All \BitResEdit{} results in this paper --- Table~\ref{tab:quantitative},
the latency in Table~\ref{tab:latency}, the per-category curves in
Figure~\ref{fig:per-category-radar}, and the reference rows of
\S\ref{sec:ablation} --- use one configuration on the released
\textsc{Infinity-2B} stack: the 2B transformer, the $D{=}32$-bit BSQ
tokenizer/VAE, and the Flan-T5-XL text encoder, run training-free in
bfloat16 with the $K{=}13$-scale 1:1 schedule ($1{\times}1$ up to
$64{\times}64$ token grids). Source images are loaded at $512^2$ and
tokenized by the source pass; edited outputs are saved at $512^2$ and
scored on all $700$ images by the standard PIE-Bench
evaluator~\citep{ju2024pnp} at $512^2$. In the notation of
\S\ref{sec:method}: CFG scale $s{=}5$; \BitEdit{} base guidance
$\eta_{0}{=}3$ with the linear anneal of \S\ref{sec:bitedit};
per-bit trust region $\delta_{\mathrm{KL}}{=}1.0$ solved with $N{=}4$
bisection steps, plus a secondary $\ell_\infty$ clamp of $7.0$ on the
deviation of the guided log-odds from the post-CFG anchor; the
spatial-bit mask $M^{k}$ is the binary edit-region gate derived from
the same localization mask, so guidance acts only inside the edit
region; sampling temperature $\tau{=}1.0$ with \textsc{Infinity}'s
default truncation (top-$k$ $900$, top-$p$ $0.97$); the unconditional
branch $\varnothing$ is \textsc{Infinity}'s learned unconditional
embedding. For \ResEdit{}, the residual strength is $\alpha{=}1.0$,
and $G_{\mathrm{final}}$ processes the benchmark's ground-truth mask
(decoded at $512^2$, white${}={}$edit) by $9{\times}9$ max-pool
dilation, bilinear resize to the final $64{\times}64$ latent grid,
$5{\times}5$ Gaussian blur ($\sigma{=}2$), and peak normalization.
The 3-branch forward (Eq.~\eqref{eq:three_branch}) runs as one
condition-batched forward at the shared visual prefix; the per-branch
hidden states and self-attention K/V caches still differ, because the
text condition enters through cross-attention and adaptive
normalization. Its FLOPs grow linearly in the number of branches but
its wall-clock cost is sub-linear (see \emph{Inference latency}
below); the KL bisection costs $N$ scalar steps per bit, and the
\ResEdit{} update is element-wise on $C\!\times\!H_z\!\times\!W_z$
tensors.
Following the PIE-Bench protocol, the source/target prompts are the
original/editing prompts (square brackets stripped). All $700$ images
use one fixed seed ($42$); repeated runs agree to within the
$\pm0.04$ CLIP-T seed-noise floor (footnote of
\S\ref{sec:ablation:components}). Latency in
Table~\ref{tab:latency} is total wall-clock (including per-worker
model loading) times the number of GPUs over $700$ images: $1821$\,s
on $4$ NVIDIA A100-80GB gives $10.41$\,s per image.

\paragraph{AREdit.}
The AREdit per-category curves in Figure~\ref{fig:per-category-radar},
its latency in Table~\ref{tab:latency}, and its column in
Figure~\ref{fig:qualitative-comparison} come from our
re-implementation of \citet{wang2025aredit} on the same
\textsc{Infinity-2B} stack (checkpoints, text encoder, $512^2$ I/O,
scale schedule, and evaluator as above); per the footnote of
Table~\ref{tab:quantitative}, the AREdit row of that table quotes the
original paper. Following the paper, the source pass tokenizes the
source image and caches its per-scale bit tokens; the edit pass
samples each scale under target-prompt CFG (scale $3.0$, temperature
$1.0$, top-$k$ $900$, top-$p$ $0.97$, guidance applied at the output
logits) and reassembles tokens at bit granularity: positions inside
the edit mask take the sampled target bits, positions outside keep
the cached source bits. We follow the paper's two-regime protocol:
for \emph{change color}, \emph{change background}, and \emph{change
style}, the cross-attention refinement controller is enabled (source
attention maps are captured and re-injected at scales up to
$16{\times}16$ token grids) with no force-preserved coarse scales
($\gamma{=}0$); for the remaining categories the controller is off
and the first $\gamma{=}3$ scales are fully preserved. Under the
GT-mask protocol shared with the other methods, the benchmark mask
(decoded at $512^2$) replaces the paper's adaptive divergence mask:
it is bilinearly resized to each scale's token grid, re-binarized at
$0.5$, and broadcast over the $32$ bits. The run covers all $700$
images with seed $42$. Latency follows the same protocol as the
other rows of Table~\ref{tab:latency}: $2072$\,s of wall-clock on
$4$ NVIDIA A100-80GB gives $11.84$\,s per image.

\paragraph{VAREdit-8B and EditInfinity.}
The VAREdit-8B and EditInfinity per-category curves in
Figure~\ref{fig:per-category-radar}, their latencies in
Table~\ref{tab:latency}, and their columns in
Figure~\ref{fig:qualitative-comparison} come from our runs of the
authors' released implementations on the full $700$-example set,
scored by the standard PIE-Bench evaluator~\citep{ju2024pnp} at
$512^2$. VAREdit-8B~\citep{mao2025varedit} uses the official
\texttt{8B-1024} checkpoint: single-pass instruction-conditioned
editing at $1024^2$ (outputs downscaled to $512^2$ for evaluation)
with CFG scale $4.0$, sampling temperature $\tau{=}1.0$, and seed
$42$. The model takes the editing instruction only --- it accepts no
localization mask --- so the GT mask enters only the evaluation.
EditInfinity~\citep{wang2025editinfinity} runs the official per-image
pipeline on the same \textsc{Infinity-2B} stack as \BitResEdit{},
with the released three-stage recipe for every image: text-embedding
optimization ($10$ iterations), LoRA fine-tuning ($50$ iterations),
and mask-blended inference, where the localization mask is the
benchmark's GT mask (decoded at $512^2$ and inverted to the
repository's white${}={}$preserve convention) --- the same GT-mask
protocol as the other mask-based methods.

\paragraph{FlowChef.}
The FlowChef latency in Table~\ref{tab:latency}, its per-category
curves in Figure~\ref{fig:per-category-radar}, and its column in
Figure~\ref{fig:qualitative-comparison} come from our run of the
authors' released implementation~\citep{Patel_2025_ICCV} on the full
$700$-example PIE-Bench set. The released editing script loads
FLUX.1-schnell in fp16 and edits at $1024\times1024$; we use the
repository's recommended editing recipe unchanged: $30$ inference
steps with steering enabled at every step, learning rate $0.6$, $10$
source steps, $5$ optimization steps per sampling step, true
classifier-free guidance $4.5$ with the source prompt as the negative
prompt, and no fixed seed. Following the PIE-Bench protocol, the
target prompt is the editing prompt, the source prompt is the
original prompt (square brackets stripped), and the localization mask
is the benchmark's ground-truth mask, decoded at $512^2$
(white${}={}$edit) and nearest-resized to $1024^2$---the same GT-mask
protocol as the other methods in
Figure~\ref{fig:per-category-radar}. Outputs are scored on all $700$
images by the standard PIE-Bench evaluator~\citep{ju2024pnp} at
$512^2$.

Latency is wall-clock around the editing call only (image loading and
saving excluded), with CUDA synchronization before and after, on a
single NVIDIA A100-80GB---the same protocol as the other rows of
Table~\ref{tab:latency}. The mean is $26.65$\,s per image (median
$26.64$\,s, p95 $26.72$\,s); the cost is constant across categories
because the recipe is fixed-step.

\paragraph{VAGS.}
The VAGS latency in Table~\ref{tab:latency}, its per-category curves
in Figure~\ref{fig:per-category-radar}, and its column in
Figure~\ref{fig:qualitative-comparison} come from our run of the
authors' released implementation~\citep{luo2026vags} (FlowEdit on
Stable Diffusion 3.5 Large, fp16, $512^2$) on the full $700$-example
PIE-Bench set, using the paper's hyperparameters: $N{=}50$ sampling
steps with $N_{\max}{=}33$ active editing steps, one noise draw per
step, source/target guidance $3.5/13.5$, and modulation strength
$\kappa{=}0.9$. Following the PIE-Bench protocol, the source/target
prompts are the original/editing prompts (square brackets stripped);
VAGS takes no localization mask, so the GT mask enters only the
evaluation, which uses the standard PIE-Bench
evaluator~\citep{ju2024pnp} at $512^2$ on all $700$ images.
One reproduction caveat: the repository's entry points wire the
constant-base cosine variant of the method
(\texttt{FlowEditSD3\_ConflictAware\_Cosine}), which in our
environment edits substantially more aggressively than the paper's
table (PSNR $23.79$, LPIPS$_{10^3}$ $97.10$, MSE$_{10^4}$ $61.89$,
SSIM$_{10^2}$ $84.64$ over the full set), even though the same
environment reproduces the paper's FlowEdit baseline row almost
exactly with $\kappa{=}0$. The released-but-unwired variant
\texttt{FlowEditSD3\_VAGS\_Cosine\_Monotone}---the same
cosine-similarity modulation applied to a monotone-increasing base
scale $\lambda_{\mathrm{tar}}(\tau)=13.5\,(0.5+0.25\,(1-\cos\pi\tau))$
over the active window---reproduces every reported metric to within
$2$ points in the table's units (PSNR $26.65$ vs.\ $26.38$, LPIPS$_{10^3}$ $68.55$ vs.\
$70.38$, MSE$_{10^4}$ $33.25$ vs.\ $34.86$, SSIM$_{10^2}$ $87.88$
vs.\ $87.68$, CLIP-Whole $26.94$ vs.\ $26.92$, CLIP-Edited $23.05$
vs.\ $23.08$); all our VAGS measurements therefore use that variant.
Latency follows the same protocol as the other rows of
Table~\ref{tab:latency}: wall-clock around the editing call only (VAE
encode, flow integration, VAE decode; loading and saving excluded),
CUDA-synchronized, on a single NVIDIA A100-80GB. The mean is
$8.65$\,s per image (median $8.64$\,s, p90 $8.70$\,s); the cost is
constant across categories because the recipe is fixed-step.

\paragraph{RewardFlow.}
The RewardFlow latency in Table~\ref{tab:latency} and its column in
Figure~\ref{fig:qualitative-comparison} come from our run
of the authors' officially released checkpoint and
pipeline~\citep{susladkar2026rewardflow} (bf16, $512^2$) on the full
$700$-example PIE-Bench set, at the editing configuration that
matches its reported PIE-Bench metrics in our reproduction: $28$
inference steps with instruction conditioning, the released SigLIP
and CLIP reward models scored against the target caption with three
reward-gradient updates per step from step $8$ onward, the reward
gradient restricted to the benchmark's GT edit mask (the same GT-mask
protocol as the other methods), an identity tether to the source
latent, and all models GPU-resident (no CPU offload). Latency is
per-image wall-clock around the editing call (per-image text
encoding, in-context flow integration, reward-gradient updates, and
VAE decode; model loading excluded), measured on single NVIDIA
A100-80GB GPUs --- the same protocol as the other rows of
Table~\ref{tab:latency}. The mean over all $700$ images is
$18.69$\,s per image (per-shard means $18.53$--$18.78$\,s across
eight category-balanced shards); the cost is constant across
categories because the recipe is fixed-step.

\paragraph{Note on the FlowChef row of Table~\ref{tab:quantitative}.}
Following the footnote of Table~\ref{tab:quantitative}, the FlowChef
row quotes reported numbers; they originate from the evaluation
in~\citet{susladkar2026rewardflow}, as the FlowChef paper itself does
not report PIE-Bench preservation metrics. Our reproduction with the
official recipe obtains comparable edit-region alignment (CLIP-Edited
$22.87$ vs.\ the quoted $23.09$) but lower background preservation
(PSNR $25.18$, LPIPS$_{10^3}$ $144.0$, MSE$_{10^4}$ $43.5$,
SSIM$_{10^2}$ $79.8$, CLIP-Whole $25.97$), in line with an
independent reproduction of FlowChef on the same
benchmark~\citep{beaudouin2025drfs}. The per-category, qualitative,
and latency results in this paper therefore use our reproduction.

\paragraph{Full-set results of our baseline runs.}
Table~\ref{tab:baseline-repro} reports the aggregate PIE-Bench
metrics of the six baseline runs described in this appendix --- the
runs behind the per-category curves of
Figure~\ref{fig:per-category-radar}, the latencies of
Table~\ref{tab:latency}, and the qualitative columns of
Figure~\ref{fig:qualitative-comparison} --- complementing
Table~\ref{tab:quantitative}, which quotes the original papers. The
two tables agree closely where the protocols coincide: EditInfinity
reproduces its quoted SSIM to within $0.01$ ($92.13$ vs.\ $92.12$)
and its CLIP scores to within at most $0.20$ (with mildly weaker
PSNR, LPIPS and MSE); VAGS
reproduces every quoted metric within $2$ points in the table's units
(see the VAGS paragraph above); and RewardFlow matches its quoted
CLIP scores within $0.2$, landing on the stronger side of its quoted
preservation numbers (e.g.\ SSIM $94.46$ vs.\ $90.21$), consistent
with the GT-masked reward gradient and identity tether of the
configuration above. Larger gaps trace to protocol differences
detailed in the per-method paragraphs: our AREdit re-implementation
replaces the paper's adaptive divergence mask with the shared GT
mask, which raises background preservation (SSIM $92.95$ vs.\ the
quoted $83.70$) and lowers text alignment; our FlowChef reproduction
preserves background less well than the numbers quoted
from~\citet{susladkar2026rewardflow} (see the preceding note); and
VAREdit-8B scores $0.65$--$1.15$ CLIP points below its quoted row
under this paper's fixed $512^2$ evaluator, while filling in the
preservation metrics its paper does not report.

\begin{table}[h]
\centering
\caption{PIE-Bench results of the baseline runs we measure ourselves,
with the \BitResEdit{} run of Table~\ref{tab:quantitative} repeated
for reference. Configurations are as described in this appendix; all
runs cover the full $700$-example set and are scored by the standard
PIE-Bench evaluator at $512^2$ (following its convention, the
background-preservation columns average over the $556$ examples
whose GT mask leaves a non-empty unedited region; the CLIP columns
over all $700$). AREdit, EditInfinity, FlowChef, and
RewardFlow use GT-mask localization; VAREdit-8B (instruction-only)
and VAGS (prompt-pair-only) take no mask. Best results are
\textbf{bold}; second-best are \underline{underlined}.}
\label{tab:baseline-repro}
\setlength{\tabcolsep}{4pt}
\renewcommand{\arraystretch}{1.1}
\sisetup{table-format=2.2, mode=text, detect-weight=true, detect-family=true}
\resizebox{\textwidth}{!}{%
\begin{tabular}{l l S S[table-format=3.2] S[table-format=3.2] S S S}
\toprule
& & \multicolumn{4}{c}{Background Preservation} & \multicolumn{2}{c}{Text Alignment} \\
\cmidrule(lr){3-6} \cmidrule(lr){7-8}
{Method} & {Backbone} & {PSNR$\uparrow$} & {LPIPS$_{10^3}\downarrow$} & {MSE$_{10^4}\downarrow$} & {SSIM$_{10^2}\uparrow$} & {Whole$\uparrow$} & {Edited$\uparrow$} \\
\midrule
VAGS~\citep{luo2026vags}                  & SD3.5-L     & 26.65 &  68.55 & {\hphantom{0}33.25} & 87.88 & 26.94 & 23.05 \\
FlowChef~\citep{Patel_2025_ICCV}          & FLUX        & 25.18 & 144.01 & {\hphantom{0}43.47} & 79.77 & 25.97 & 22.87 \\
RewardFlow~\citep{susladkar2026rewardflow} & Qwen-Image-20B & \bfseries 31.00 & {\hphantom{0}\bfseries 27.07} & {\hphantom{0}\bfseries 13.86} & \bfseries 94.46 & \bfseries 29.61 & \bfseries 27.45 \\
\midrule
AREdit~\citep{wang2025aredit}             & Infinity-2B & {\underline{29.82}} & {\hphantom{0}\underline{34.83}} & {\hphantom{0}\underline{21.80}} & {\underline{92.95}} & 24.26 & 21.30 \\
VAREdit-8B~\citep{mao2025varedit}         & Infinity-8B & 26.54 &  67.16 & {\hphantom{0}73.77} & 89.10 & 25.45 & 22.65 \\
EditInfinity~\citep{wang2025editinfinity} & Infinity-2B & 27.41 &  40.11 & {\hphantom{0}28.38} & 92.13 & 26.36 & 23.67 \\
\textbf{BitResEdit (ours)}                & Infinity-2B & 26.67 &  43.21 & {\hphantom{0}31.61} & 91.71 & {\underline{27.13}} & {\underline{24.54}} \\
\bottomrule
\end{tabular}%
}
\end{table}

\paragraph{Inference latency.}
Table~\ref{tab:latency} reports per-image wall-clock latency averaged
over the full $700$-example PIE-Bench test set on a single NVIDIA
A100-80GB GPU. Per-image fine-tuning makes EditInfinity
$\sim\!45\times$ slower than \BitResEdit{} ($466.77$\,s vs
$10.41$\,s, i.e.\ about $7.8$ minutes vs $10$ seconds per image),
while the three training-free VAR methods differ by under $8$ seconds. VAREdit-8B is the fastest at $4.70$\,s despite its larger
Infinity-8B backbone, because it issues a single forward pass per
image; \BitResEdit{} runs its $3$-branch \BitEdit{} guidance and the
\ResEdit{} residual injection in $10.41$\,s, comparable to AREdit's
$11.84$\,s on the same Infinity-2B backbone. Both methods run the
same two autoregressive passes (source caching and editing), and at a
per-scale batch of $2$--$3$ the transformer forward is dominated by
weight loading, so the third guidance branch of \BitResEdit{} adds
little. AREdit's extra $1.4$\,s comes from its attention refinement:
capturing source cross-attention maps and re-injecting them during
editing materializes attention weights explicitly and bypasses the
fused attention kernel. \BitResEdit{} acts only on output log-odds
and residual codes, leaving every attention layer on the fused path.
The FLUX-based
FlowChef, measured with its official $30$-step editing recipe
(described above), takes $26.65$\,s per image, about
$2.6\times$ \BitResEdit{}: each sampling step evaluates the FLUX
transformer twice for classifier-free guidance and adds
gradient-steering updates on top. RewardFlow, measured with its
officially released checkpoint and reward-guided pipeline
(described above), takes $18.69$\,s per image, about
$1.8\times$ \BitResEdit{}: its in-context source conditioning doubles
the transformer's token sequence, and each reward-guided step adds
gradient-ascent updates that backpropagate through the VAE decoder
and its CLIP and SigLIP reward models. The SD3.5-based VAGS, measured
with the configuration that reproduces its reported results
(described above), is the second-fastest method at
$8.65$\,s: its $33$ active flow-matching steps each evaluate the
SD3.5 transformer once on a four-way batch, and the adaptive
guidance scale adds no extra forward passes.

\begin{table}[h]
\centering
\caption{PIE-Bench inference latency. Per-image wall-clock is
averaged over the full $700$-example test set on an NVIDIA A100-80GB
GPU. EditInfinity latency includes per-image fine-tuning (the
user-facing cost); the other methods are training-free. FlowChef,
VAGS, and RewardFlow are our measurements of the official editing
pipelines (described in this appendix).}
\label{tab:latency}
\setlength{\tabcolsep}{6pt}
\renewcommand{\arraystretch}{1.1}
\sisetup{table-format=3.2, mode=text, detect-weight=true, detect-family=true}
\begin{tabular}{l l S}
\toprule
{Method} & {Backbone} & {Latency (s/img)$\downarrow$} \\
\midrule
FlowChef~\citep{Patel_2025_ICCV}          & FLUX        & {\hphantom{0}26.65} \\
VAGS~\citep{luo2026vags}                  & SD3.5-L     & {\hphantom{00}\underline{8.65}} \\
RewardFlow~\citep{susladkar2026rewardflow} & Qwen-Image-20B & {\hphantom{0}18.69} \\
\midrule
EditInfinity~\citep{wang2025editinfinity} & Infinity-2B & 466.77 \\
AREdit~\citep{wang2025aredit}             & Infinity-2B &  11.84 \\
VAREdit-8B~\citep{mao2025varedit}         & Infinity-8B & {\hphantom{00}\bfseries 4.70} \\
\textbf{BitResEdit (ours)}                & Infinity-2B & {\hphantom{0}10.41} \\
\bottomrule
\end{tabular}
\end{table}

%% file: chapters/pseudocode.tex
% ---------------------------------------------------------------------------
% Appendix: PyTorch-like pseudocode for BitResEdit, in the terse "Algorithm N"
% style of Kaiming He's recent papers (MeanFlow, improved MeanFlow, JiT).
% The ruled "kalg" box and the Kaiming-style listing colors are defined in
% new_commands.tex (shared with Algorithm 1, the edit pass, which appears in
% the method section); only \usepackage{listings} is needed in each preamble.
% ---------------------------------------------------------------------------

\section{PyTorch-Style Pseudocode for \BitResEdit{}}
\label{sec:pseudocode}

Algorithm~\ref{alg:edit} (the edit pass, \S\ref{sec:bitresedit}) and
Algorithm~\ref{alg:klclip} below (the Bernoulli-KL trust-region projection)
give a PyTorch-like transcription of
\BitResEdit{} (\S\ref{sec:method}), in the terse style of recent
flow-based work; code indices run $0\,..\,K\!-\!1$. At each of the $K\!=\!13$
scales, \BitEdit{} runs a 3-branch forward (Eq.~\ref{eq:three_branch}), forms
the post-CFG per-bit log-odds (Eq.~\ref{eq:cfg}), tilts them along the
source-negative direction under the edit-region bit gate
(Eqs.~\ref{eq:neg_direction},~\ref{eq:bitedit_raw}), and projects the result
onto a per-bit Bernoulli-KL ball around the post-temperature CFG anchor
(Eqs.~\ref{eq:bitedit_proj}--\ref{eq:bitedit_alpha}), followed by the
element-wise clamp of \S\ref{sec:bitedit}; \ResEdit{} writes the
mask-gated code residual at the final latent resolution (Eq.~\ref{eq:resedit})
and accumulates it into the prefix carried to the next scale. Defaults
(Appendix~\ref{app:impl-details}): $\eta_{k}$ linearly annealed from
$\eta_{0}\!=\!3$ to $0$, $\delta_{\mathrm{KL}}\!=\!1.0$, clamp
$\mathrm{lim}\!=\!7$, $\alpha\!=\!1$, $N\!=\!4$ bisection steps; the CFG scale
$s$, temperature $\tau$, and top-$k$/top-$p$ truncation
follow our \textsc{Infinity}~\citep{han2025infinity} setup. Code-space tensors
are $C\!\times\!H_z\!\times\!W_z$; per-bit log-odds are $P_k\!\times\!D$.

\begin{kalg}{\BitEdit{}: Bernoulli-KL trust-region projection \texttt{kl\_clip}}{\texttt{bern\_kl} increases along the segment from anchor \texttt{a} to \texttt{d\_raw}, so the bracket tracks a feasible lower bound.}
\label{alg:klclip}
\begin{lstlisting}[style=kaiming]
# kl_clip(d_raw, a, delta): pull d_raw toward anchor a until
#   its per bit Bernoulli KL to a is within delta, N steps
# bern_kl(p, q): Bernoulli KL from Bern(p) to Bern(q)

lo, hi = zeros_like(d_raw), ones_like(d_raw)  # mix weight bracket
for _ in range(N):
    m = (lo + hi) / 2
    pm = sigmoid(m * d_raw + (1 - m) * a)
    ok = bern_kl(pm, sigmoid(a)) <= delta     # inside the ball?
    lo = where(ok, m, lo)                      # largest feasible m
    hi = where(ok, hi, m)
full = bern_kl(sigmoid(d_raw), sigmoid(a)) <= delta
lam = where(full, ones_like(lo), lo)          # full update
return lam * d_raw + (1 - lam) * a
\end{lstlisting}
\end{kalg}

%% file: chapters/limitations.tex
% Limitations section: discusses the dependence of BitResEdit on the
% quality of the input localization mask, citing the mask-robustness
% ablation (Table~\ref{tab:ablation:mask}) for empirical magnitudes.
\section{Limitations}
\label{sec:limitations}

\BitResEdit{} assumes the user supplies a binary localization mask
$M_{\mathrm{gt}}$ that delineates the region to be edited. Both
algebraic sites in our editor read this mask: \ResEdit{}'s mask-gated
sum-of-scales sets the residual to exactly zero at masked-out
positions, and \BitEdit{}'s per-bit Bernoulli-KL trust region is
applied through the same gate $G_{\mathrm{final}}$. Mask quality
therefore lies directly on the path from prompt to edit, and our
guarantees on background preservation and on bounded drift from the
clean CFG sampler are guarantees \emph{relative to the supplied mask},
not relative to the semantic region a user has in mind.

The mask-robustness ablation in
\S\ref{sec:ablation:mask} (Table~\ref{tab:ablation:mask}) makes
this dependence explicit. Replacing the PIE-Bench ground-truth mask
with a tight bounding box of the same region drops PSNR from $26.67$
to $19.68$\,dB and roughly triples LPIPS$_{10^{3}}$ ($43.21\!\to\!130.07$),
because edits now spread across the entire bounding rectangle even
where the source content should have been preserved. An automatic
GroundingDINO+SAM pipeline closes most of this gap ($25.75$\,dB PSNR,
$47.82$ LPIPS$_{10^{3}}$) but still trails the ground-truth mask, and
inherits the failure modes of its grounding and segmentation
components: missed or hallucinated entities, ambiguous referring
expressions, occluded objects, and fine-structured or thin boundaries.
For deployments without ground-truth masks, \BitResEdit{} therefore
inherits the reliability of the upstream localization pipeline, and a
poor mask can suppress wanted edits inside the region or leak edits
into the background even when \BitEdit{} and \ResEdit{} themselves
behave as designed. Jointly inferring or refining the mask together
with the per-bit edit, or deriving a localization signal directly from
the per-bit Bernoulli geometry that \BitResEdit{} already computes,
are natural directions for future work.

%% file: chapters/failure_cases.tex
% Failure-case study for the object-deletion category. Presents three
% representative PIE-Bench deletion failures of BitResEdit (ground-truth mask,
% v1_x1_x0_x3 forward run) and explains why a residual editor underperforms on
% removal. Numbers are per-category means over the 80 PIE-Bench deletion images.
% \promptdel is also defined in intro.tex; \providecommand keeps this file safe
% if it is ever compiled without the intro.
\providecommand{\promptdel}[1]{{\color{red}#1}}

\section{Failure Cases: Object Deletion}
\label{sec:failure-deletion}

\begin{figure}[t]
  \centering
  \newcommand{\failw}{0.31\linewidth}
  \makebox[\failw]{\footnotesize Source}\hfill
  \makebox[\failw]{\footnotesize GT mask (edit region)}\hfill
  \makebox[\failw]{\footnotesize \BitResEdit{}}\\[2pt]
  % (a) salient object recoloured rather than removed.
  \includegraphics[width=\failw]{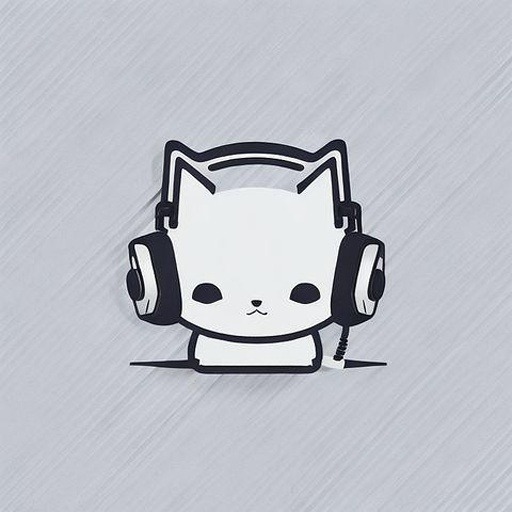}\hfill
  \includegraphics[width=\failw]{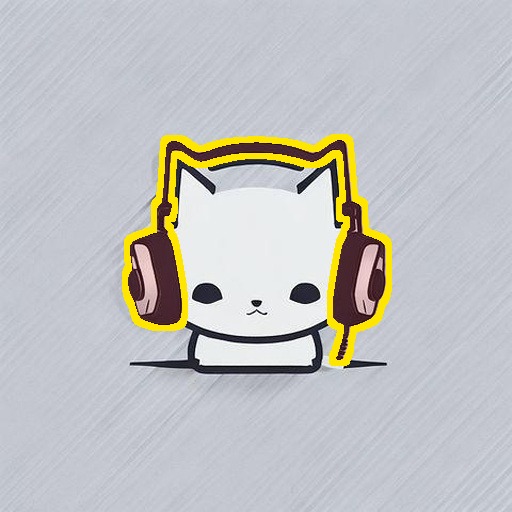}\hfill
  \includegraphics[width=\failw]{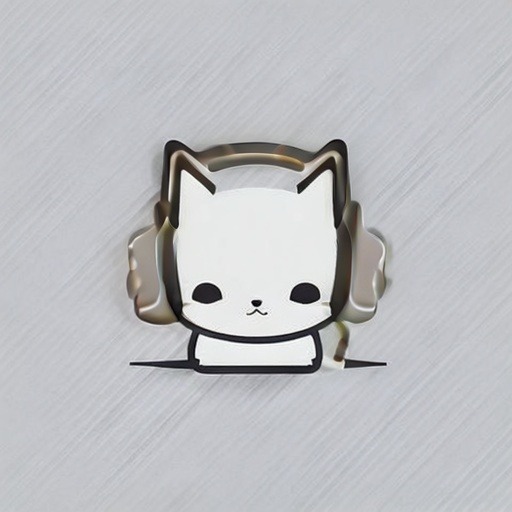}\\[1pt]
  {\scriptsize (a)~``a cat \promptdel{wearing headphones} on a gray background.''
  The masked headphones are recoloured into a washed-out form, not removed.\par}\smallskip
  % (b) emptied region filled with incoherent content.
  \includegraphics[width=\failw]{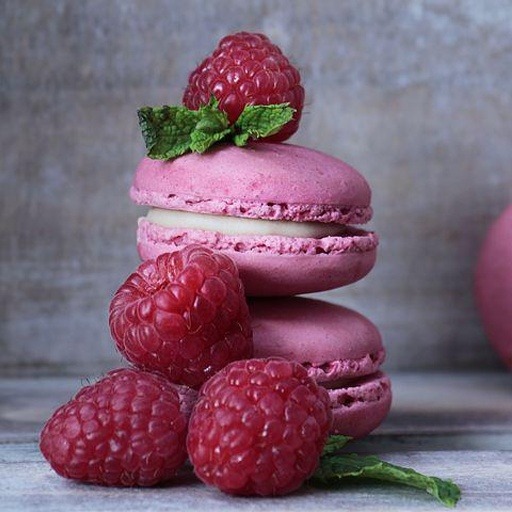}\hfill
  \includegraphics[width=\failw]{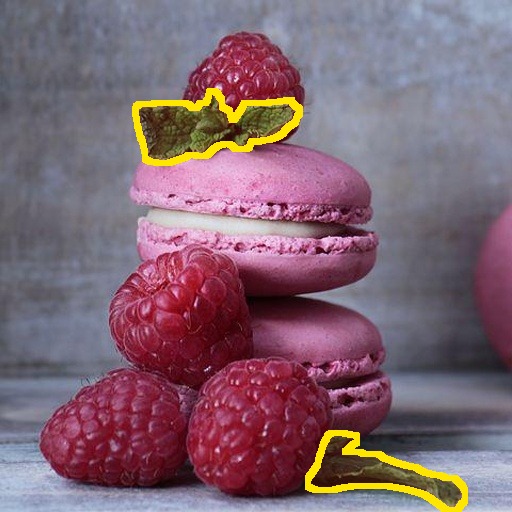}\hfill
  \includegraphics[width=\failw]{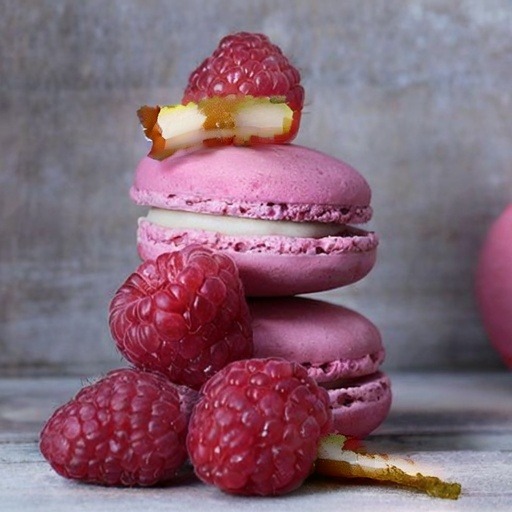}\\[1pt]
  {\scriptsize (b)~``raspberry macarons \promptdel{with mint leaves}.'' The mint
  is replaced by an incoherent fruit-like smear (the category's lowest
  $\text{CLIP}_{\text{edit}}$, $8.6$).\par}\smallskip
  % (c) diffuse target forces near-global repainting.
  \includegraphics[width=\failw]{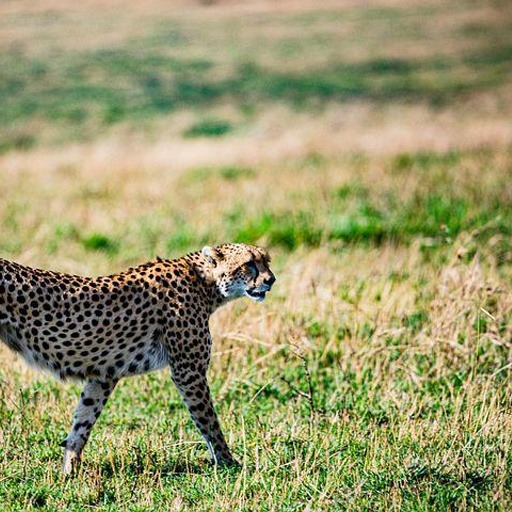}\hfill
  \includegraphics[width=\failw]{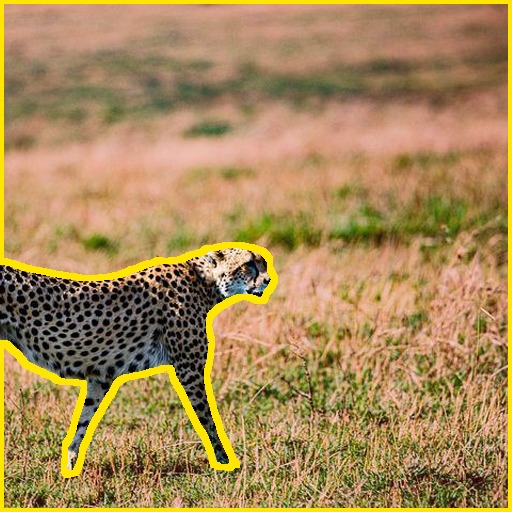}\hfill
  \includegraphics[width=\failw]{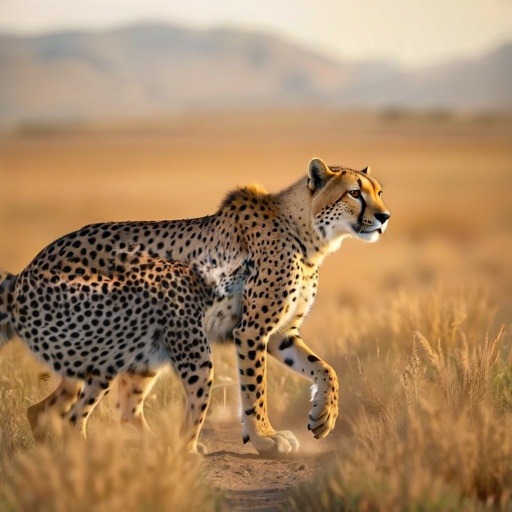}\\[1pt]
  {\scriptsize (c)~``a cheetah walking across a \promptdel{grassy} field.'' The
  diffuse target makes the ground-truth mask span $88\%$ of the image; the field
  is repainted and a spurious second cheetah is hallucinated, while only the
  masked-out animal is preserved.\par}
  \caption{Representative object-deletion failures of \BitResEdit{} on PIE-Bench
  under ground-truth masks. Columns: source, the supplied edit region (red fill,
  yellow outline), and the \BitResEdit{} output. The rows show the three dominant
  failure modes: (a)~a salient object is restyled rather than removed; (b)~the
  emptied region is filled with incoherent content; (c)~a diffuse target forces
  near-global repainting that corrupts preserved content. Red words mark the
  deleted phrase.}
  \label{fig:failure-deletion}
\end{figure}

Object deletion is the category on which \BitResEdit{} aligns least well with
the edit instruction. Under the ground-truth mask it reaches a mean edit-region
alignment of $\text{CLIP}_{\text{edit}}=19.4$, the lowest of all ten PIE-Bench
editing types and well below the $24.5$ macro-average
(Figure~\ref{fig:per-category-radar}), even though the same images retain strong
background preservation---mean structure distance $0.023$ and PSNR $27.0$\,dB,
among the strongest in the benchmark. This combination---faithful preservation
but weak edited content---is intrinsic to how a residual editor handles removal.
Figure~\ref{fig:failure-deletion} shows three representative cases.

\paragraph{Why deletion is hard for a residual editor.}
\BitResEdit{} is preservation-biased by construction: \ResEdit{} sets the
residual to exactly zero outside the mask, and inside the mask \BitEdit{} only
tilts the post-CFG log-odds toward the target prompt within a per-bit
Bernoulli-KL trust region around the clean sampler. For most edits this is
exactly right---the target prompt positively names the new content (a different
object, colour, or style), and the editor moves the masked bits toward it.
Deletion is different: the target prompt is \emph{subtractive}, naming what
should disappear but nothing that should take its place. \textsc{Infinity} is a
generation prior, not an inpainting prior; it is never asked to continue the
surrounding background, so inside the mask it must synthesise content from a
prompt that specifies none. The three modes in Figure~\ref{fig:failure-deletion}
follow directly. (a)~When the masked object is salient, source-negative guidance
restyles it instead of erasing it---the coarse-scale structure of a strong object
survives the trust-region nudge. (b)~Once the object is gone, the under-specified
fill collapses to incoherent content rather than clean background. (c)~When the
deletion target is a diffuse texture such as ``grass,'' the \emph{correct}
ground-truth mask already covers most of the image, so the editor must repaint
almost the entire scene with no positive target, recolouring it and hallucinating
spurious objects.

\paragraph{The metric overstates the deficit.}
Part of the low $\text{CLIP}_{\text{edit}}$ is structural rather than a true
error. $\text{CLIP}_{\text{edit}}$ scores the agreement between the edited region
and the target caption, but a successful deletion turns that region into
background, which has little positive text to align with, so the score is
depressed even when removal succeeds. The $19.4$ figure therefore conflates
genuine failures with an inherent mismatch between deletion and a text-alignment
metric---consistent with deletion's far healthier whole-image CLIP of $25.2$.
The visible failures in Figure~\ref{fig:failure-deletion} are nonetheless real,
and they are distinct from the mask-quality limitations of
Section~\ref{sec:limitations}: here the masks are the correct ground-truth
regions, and the difficulty lies in synthesising background for an
under-determined region, not in localisation. A natural remedy is to give the
masked region an explicit inpainting target---for example, conditioning the
in-mask generation on the surrounding context or compositing against a background
estimate---rather than relying on a subtractive prompt alone.

%% file: chapters/llm_usage.tex
% Declaration of LLM usage. Required by NeurIPS 2026 when LLMs are used
% in any capacity that warrants disclosure; we list the three concrete
% roles LLMs played during the preparation of this work.
\section{Declaration of LLM Usage}
\label{sec:llmusage}

We used large language models (LLMs) as assistive tools during the
preparation of this work, in three roles: (i) as coding assistants
when writing experiment, evaluation, and plotting scripts; (ii) as a
sounding board when brainstorming design alternatives for
\BitResEdit{}; and (iii) for clarifying background concepts and prior
work. All LLM-generated code and any technical claim attributed to
such discussions were reviewed and verified by the authors before
inclusion. LLMs were not used to generate or alter any experimental
results, and they are not part of the core methodology of
\BitResEdit{} or of any baseline.

%% file: literature.bib
@inproceedings{oord2017neural,
  title     = {Neural Discrete Representation Learning},
  author    = {van den Oord, Aaron and Vinyals, Oriol and Kavukcuoglu, Koray},
  booktitle = {Advances in Neural Information Processing Systems},
  year      = {2017}
}

@inproceedings{chen2020generative,
  title     = {Generative Pretraining From Pixels},
  author    = {Chen, Mark and Radford, Alec and Child, Rewon and Wu, Jeffrey and Jun, Heewoo and Luan, David and Sutskever, Ilya},
  booktitle = {Proceedings of the 37th International Conference on Machine Learning},
  pages     = {1691--1703},
  year      = {2020},
  volume    = {119},
  series    = {Proceedings of Machine Learning Research},
  publisher = {PMLR}
}

@inproceedings{esser2021taming,
  title     = {Taming Transformers for High-Resolution Image Synthesis},
  author    = {Esser, Patrick and Rombach, Robin and Ommer, Bj{\"o}rn},
  booktitle = {Proceedings of the IEEE/CVF Conference on Computer Vision and Pattern Recognition},
  year      = {2021}
}

@inproceedings{ramesh2021zero,
  title     = {Zero-Shot Text-to-Image Generation},
  author    = {Ramesh, Aditya and Pavlov, Mikhail and Goh, Gabriel and Gray, Scott and Voss, Chelsea and Radford, Alec and Chen, Mark and Sutskever, Ilya},
  booktitle = {Proceedings of the 38th International Conference on Machine Learning},
  year      = {2021}
}

@article{yu2022scaling,
  title   = {Scaling Autoregressive Models for Content-Rich Text-to-Image Generation},
  author  = {Yu, Jiahui and Xu, Yuanzhong and Koh, Jing Yu and Luong, Thang and Baid, Gunjan and Wang, Zirui and Vasudevan, Vijay and Ku, Alexander and Yang, Yinfei and Ayan, Burcu Karagol and Hutchinson, Ben and Han, Wei and Parekh, Zarana and Li, Xin and Zhang, Han and Baldridge, Jason and Wu, Yonghui},
  journal = {arXiv preprint arXiv:2206.10789},
  year    = {2022}
}

@inproceedings{tian2024visual,
  title     = {Visual Autoregressive Modeling: Scalable Image Generation via Next-Scale Prediction},
  author    = {Tian, Keyu and Jiang, Yi and Yuan, Zehuan and Peng, Bingyue and Wang, Liwei},
  booktitle = {Advances in Neural Information Processing Systems},
  year      = {2024}
}

@article{sun2024autoregressive,
  title   = {Autoregressive Model Beats Diffusion: Llama for Scalable Image Generation},
  author  = {Sun, Peize and Jiang, Yi and Chen, Shoufa and Zhang, Shilong and Peng, Bingyue and Luo, Ping and Yuan, Zehuan},
  journal = {arXiv preprint arXiv:2406.06525},
  year    = {2024}
}

@article{ren2024mvar,
  title   = {M-VAR: Decoupled Scale-wise Autoregressive Modeling for High-Quality Image Generation},
  author  = {Ren, Sucheng and Yu, Yaodong and Ruiz, Nataniel and Wang, Feng and Yuille, Alan and Xie, Cihang},
  journal = {arXiv preprint arXiv:2411.10433},
  year    = {2024}
}

@article{ren2024flowar,
  title   = {FlowAR: Scale-wise Autoregressive Image Generation Meets Flow Matching},
  author  = {Ren, Sucheng and Yu, Qihang and He, Ju and Shen, Xiaohui and Yuille, Alan and Chen, Liang-Chieh},
  journal = {arXiv preprint arXiv:2412.15205},
  year    = {2024}
}

@inproceedings{han2025infinity,
  title     = {Infinity: Scaling Bitwise AutoRegressive Modeling for High-Resolution Image Synthesis},
  author    = {Han, Jian and Liu, Jinlai and Jiang, Yi and Yan, Bin and Zhang, Yuqi and Yuan, Zehuan and Peng, Bingyue and Liu, Xiaobing},
  booktitle = {Proceedings of the IEEE/CVF Conference on Computer Vision and Pattern Recognition},
  pages     = {15733--15744},
  year      = {2025}
}

@article{jiao2025flexvar,
  title   = {FlexVAR: Flexible Visual Autoregressive Modeling without Residual Prediction},
  author  = {Jiao, Siyu and Zhang, Gengwei and Qian, Yinlong and Huang, Jiancheng and Zhao, Yao and Shi, Humphrey and Ma, Lin and Wei, Yunchao and Jie, Zequn},
  journal = {arXiv preprint arXiv:2502.20313},
  year    = {2025}
}

@inproceedings{shao2025continuous,
  title     = {Continuous Visual Autoregressive Generation via Score Maximization},
  author    = {Shao, Chenze and Meng, Fandong and Zhou, Jie},
  booktitle = {Proceedings of the 42nd International Conference on Machine Learning},
  year      = {2025}
}

@inproceedings{kumbong2025hmar,
  title     = {HMAR: Efficient Hierarchical Masked Auto-Regressive Image Generation},
  author    = {Kumbong, Hermann and Liu, Xian and Lin, Tsung-Yi and Liu, Ming-Yu and Liu, Xihui and Liu, Ziwei and Fu, Daniel Y. and Re, Christopher and Romero, David W.},
  booktitle = {Proceedings of the IEEE/CVF Conference on Computer Vision and Pattern Recognition},
  year      = {2025}
}

@article{liu2025infinitystar,
  title   = {InfinityStar: Unified Spacetime AutoRegressive Modeling for Visual Generation},
  author  = {Liu, Jinlai and Han, Jian and Yan, Bin and Wu, Hui and Zhu, Fengda and Wang, Xing and Jiang, Yi and Peng, Bingyue and Yuan, Zehuan},
  journal = {arXiv preprint arXiv:2511.04675},
  year    = {2025}
}

@inproceedings{meng2021sdedit,
  title     = {SDEdit: Guided Image Synthesis and Editing with Stochastic Differential Equations},
  author    = {Meng, Chenlin and He, Yutong and Song, Yang and Song, Jiaming and Wu, Jiajun and Zhu, Jun-Yan and Ermon, Stefano},
  booktitle = {International Conference on Learning Representations},
  year      = {2022}
}

@article{hertz2022prompt,
  title   = {Prompt-to-Prompt Image Editing with Cross Attention Control},
  author  = {Hertz, Amir and Mokady, Ron and Tenenbaum, Jay and Aberman, Kfir and Pritch, Yael and Cohen-Or, Daniel},
  journal = {arXiv preprint arXiv:2208.01626},
  year    = {2022}
}

@inproceedings{mokady2023null,
  title     = {NULL-text Inversion for Editing Real Images Using Guided Diffusion Models},
  author    = {Mokady, Ron and Hertz, Amir and Aberman, Kfir and Pritch, Yael and Cohen-Or, Daniel},
  booktitle = {Proceedings of the IEEE/CVF Conference on Computer Vision and Pattern Recognition},
  year      = {2023}
}

@inproceedings{couairon2022diffedit,
  title     = {DiffEdit: Diffusion-based Semantic Image Editing with Mask Guidance},
  author    = {Couairon, Guillaume and Verbeek, Jakob and Schwenk, Holger and Cord, Matthieu},
  booktitle = {International Conference on Learning Representations},
  year      = {2023}
}

@inproceedings{tumanyan2023plug,
  title     = {Plug-and-Play Diffusion Features for Text-Driven Image-to-Image Translation},
  author    = {Tumanyan, Narek and Geyer, Michal and Bagon, Shai and Dekel, Tali},
  booktitle = {Proceedings of the IEEE/CVF Conference on Computer Vision and Pattern Recognition},
  year      = {2023}
}

@inproceedings{parmar2023zero,
  title     = {Zero-shot Image-to-Image Translation},
  author    = {Parmar, Gaurav and Singh, Krishna Kumar and Zhang, Richard and Li, Yijun and Lu, Jingwan and Zhu, Jun-Yan},
  booktitle = {ACM SIGGRAPH Conference Proceedings},
  year      = {2023}
}

@inproceedings{cao2023masactrl,
  title     = {MasaCtrl: Tuning-Free Mutual Self-Attention Control for Consistent Image Synthesis and Editing},
  author    = {Cao, Mingdeng and Wang, Xintao and Qi, Zhongang and Shan, Ying and Qie, Xiaohu and Zheng, Yinqiang},
  booktitle = {Proceedings of the IEEE/CVF International Conference on Computer Vision},
  year      = {2023}
}

@inproceedings{brooks2023instructpix2pix,
  title     = {InstructPix2Pix: Learning to Follow Image Editing Instructions},
  author    = {Brooks, Tim and Holynski, Aleksander and Efros, Alexei A.},
  booktitle = {Proceedings of the IEEE/CVF Conference on Computer Vision and Pattern Recognition},
  year      = {2023}
}

@inproceedings{mu2025editar,
  title     = {EditAR: Unified Conditional Generation with Autoregressive Models},
  author    = {Mu, Jiteng and Vasconcelos, Nuno and Wang, Xiaolong},
  booktitle = {Proceedings of the IEEE/CVF Conference on Computer Vision and Pattern Recognition},
  year      = {2025}
}

@inproceedings{wang2025aredit,
  title     = {Training-Free Text-Guided Image Editing with Visual Autoregressive Model},
  author    = {Wang, Yufei and Guo, Lanqing and Li, Zhihao and Huang, Jiaxing and Wang, Pichao and Wen, Bihan and Wang, Jian},
  booktitle = {Proceedings of the IEEE/CVF International Conference on Computer Vision},
  year      = {2025}
}

@inproceedings{mao2025varedit,
  title     = {Visual Autoregressive Modeling for Instruction-Guided Image Editing},
  author    = {Mao, Qingyang and Cai, Qi and Li, Yehao and Pan, Yingwei and Cheng, Mingyue and Yao, Ting and Liu, Qi and Mei, Tao},
  booktitle = {International Conference on Learning Representations},
  year      = {2026}
}

@inproceedings{ahmadi2025earl,
  title     = {The Promise of RL for Autoregressive Image Editing},
  author    = {Ahmadi, Saba and Awal, Rabiul and Sikarwar, Ankur and Kazemnejad, Amirhossein and Luo, Ge Ya and Rodriguez, Juan A. and Rajeswar, Sai and Reddy, Siva and Pal, Christopher and Krojer, Benno and Agrawal, Aishwarya},
  booktitle = {Advances in Neural Information Processing Systems},
  year      = {2025}
}

@article{elghoussani2026mln,
  title   = {Prompt-Guided Image Editing with Masked Logit Nudging in Visual Autoregressive Models},
  author  = {El-Ghoussani, Amir and H{\"o}lle, Marc and Carneiro, Gustavo and Belagiannis, Vasileios},
  journal = {arXiv preprint arXiv:2604.14591},
  year    = {2026}
}

@article{xia2026structure,
  title   = {Rethinking Structure Preservation in Text-Guided Image Editing with Visual Autoregressive Models},
  author  = {Xia, Tao and Liu, Jiawei and Zhang, Yukun and Liu, Ting and Wang, Wei and Zhang, Lei},
  journal = {arXiv preprint arXiv:2603.28367},
  year    = {2026}
}

@article{dao2025varin,
  title   = {Discrete Noise Inversion for Next-scale Autoregressive Text-based Image Editing},
  author  = {Dao, Quan and He, Xiaoxiao and Han, Ligong and Nguyen, Ngan Hoai and Heyrani Nobari, Amin and Ahmed, Faez and Zhang, Han and Nguyen, Viet Anh and Metaxas, Dimitris},
  journal = {arXiv preprint arXiv:2509.01984},
  year    = {2025}
}

@inproceedings{wang2025editinfinity,
  title     = {EditInfinity: Image Editing with Binary-Quantized Generative Models},
  author    = {Wang, Jiahuan and Chen, Yuxin and Yu, Jun and Lu, Guangming and Pei, Wenjie},
  booktitle = {Advances in Neural Information Processing Systems},
  year      = {2025}
}

@inproceedings{zhao2024bsq,
  title     = {Image and Video Tokenization with Binary Spherical Quantization},
  author    = {Zhao, Yue and Xiong, Yuanjun and Kr{\"a}henb{\"u}hl, Philipp},
  booktitle = {International Conference on Learning Representations},
  year      = {2025}
}

@article{ho2022classifier,
  title   = {Classifier-Free Diffusion Guidance},
  author  = {Ho, Jonathan and Salimans, Tim},
  journal = {arXiv preprint arXiv:2207.12598},
  year    = {2022}
}

@inproceedings{ju2024pnp,
  title     = {PnP Inversion: Boosting Diffusion-based Editing with 3 Lines of Code},
  author    = {Ju, Xuan and Zeng, Ailing and Bian, Yuxuan and Liu, Shaoteng and Xu, Qiang},
  booktitle = {International Conference on Learning Representations},
  year      = {2024},
  url       = {https://openreview.net/forum?id=FoMZ4ljhVw}
}

@inproceedings{Miyake_2025_WACV,
  author    = {Miyake, Daiki and Iohara, Akihiro and Saito, Yu and Tanaka, Toshiyuki},
  title     = {Negative-Prompt Inversion: Fast Image Inversion for Editing with Text-Guided Diffusion Models},
  booktitle = {Proceedings of the Winter Conference on Applications of Computer Vision (WACV)},
  month     = {February},
  year      = {2025},
  pages     = {2063--2072}
}

@inproceedings{Avrahami_2025_CVPR,
  author    = {Avrahami, Omri and Patashnik, Or and Fried, Ohad and Nemchinov, Egor and Aberman, Kfir and Lischinski, Dani and Cohen-Or, Daniel},
  title     = {Stable Flow: Vital Layers for Training-Free Image Editing},
  booktitle = {Proceedings of the Computer Vision and Pattern Recognition Conference (CVPR)},
  month     = {June},
  year      = {2025},
  pages     = {7877--7888}
}

@inproceedings{wang2025taming,
  title     = {Taming Rectified Flow for Inversion and Editing},
  author    = {Wang, Jiangshan and Pu, Junfu and Qi, Zhongang and Guo, Jiayi and Ma, Yue and Huang, Nisha and Chen, Yuxin and Li, Xiu and Shan, Ying},
  booktitle = {International Conference on Machine Learning},
  year      = {2025},
  url       = {https://openreview.net/forum?id=uDreZphNky}
}

@inproceedings{Kulikov_2025_ICCV,
  author    = {Kulikov, Vladimir and Kleiner, Matan and Huberman-Spiegelglas, Inbar and Michaeli, Tomer},
  title     = {FlowEdit: Inversion-Free Text-Based Editing Using Pre-Trained Flow Models},
  booktitle = {Proceedings of the IEEE/CVF International Conference on Computer Vision (ICCV)},
  month     = {October},
  year      = {2025},
  pages     = {19721--19730}
}

@inproceedings{Patel_2025_ICCV,
  author    = {Patel, Maitreya and Wen, Song and Metaxas, Dimitris N. and Yang, Yezhou},
  title     = {FlowChef: Steering of Rectified Flow Models for Controlled Generations},
  booktitle = {Proceedings of the IEEE/CVF International Conference on Computer Vision (ICCV)},
  month     = {October},
  year      = {2025},
  pages     = {15308--15318}
}

@misc{susladkar2026rewardflow,
  title         = {RewardFlow: Generate Images by Optimizing What You Reward},
  author        = {Susladkar, Onkar and Jang, Dong-Hwan and Prakash, Tushar and Juvekar, Adheesh and Shah, Vedant and Barik, Ayush and Bashir, Nabeel and Wahed, Muntasir and Shrirao, Ritish and Lourentzou, Ismini},
  year          = {2026},
  eprint        = {2604.08536},
  archiveprefix = {arXiv},
  primaryclass  = {cs.CV},
  note          = {CVPR 2026}
}

@inproceedings{xu2024infedit,
  title     = {Inversion-Free Image Editing with Language-Guided Diffusion Models},
  author    = {Xu, Sihan and Huang, Yidong and Pan, Jiayi and Ma, Ziqiao and Chai, Joyce},
  booktitle = {Proceedings of the IEEE/CVF Conference on Computer Vision and Pattern Recognition},
  year      = {2024}
}

@inproceedings{brack2023ledits++,
  title     = {{LEDITS++}: Limitless Image Editing using Text-to-Image Models},
  author    = {Brack, Manuel and Friedrich, Felix and Kornmeier, Katharina and Tsaban, Linoy and Schramowski, Patrick and Kersting, Kristian and Passos, Apolin{\'a}rio},
  booktitle = {Proceedings of the IEEE/CVF Conference on Computer Vision and Pattern Recognition},
  year      = {2024}
}

@article{tang2024locinv,
  title   = {{LocInv}: Localization-aware Inversion for Text-Guided Image Editing},
  author  = {Tang, Chuanming and Wang, Kai and Yang, Fei and van de Weijer, Joost},
  journal = {arXiv preprint arXiv:2405.01496},
  year    = {2024}
}

@inproceedings{sheynin2024emu,
  title     = {Emu Edit: Precise Image Editing via Recognition and Generation Tasks},
  author    = {Sheynin, Shelly and Polyak, Adam and Singer, Uriel and Kirstain, Yuval and Zohar, Amit and Ashual, Oron and Parikh, Devi and Taigman, Yaniv},
  booktitle = {Proceedings of the IEEE/CVF Conference on Computer Vision and Pattern Recognition},
  year      = {2024}
}

@inproceedings{zhao2024ultraedit,
  title     = {UltraEdit: Instruction-based Fine-Grained Image Editing at Scale},
  author    = {Zhao, Haozhe and others},
  booktitle = {Advances in Neural Information Processing Systems},
  year      = {2024}
}

@inproceedings{wei2024omniedit,
  title     = {OmniEdit: Building Image Editing Generalist Models Through Specialist Supervision},
  author    = {Wei, Cong and Xiong, Zheyang and Ren, Weiming and Du, Xinrun and Zhang, Ge and Chen, Wenhu},
  booktitle = {International Conference on Learning Representations},
  year      = {2025}
}

@inproceedings{yu2025anyedit,
  title     = {AnyEdit: Mastering Unified High-Quality Image Editing for Any Idea},
  author    = {Yu, Qifan and others},
  booktitle = {Proceedings of the IEEE/CVF Conference on Computer Vision and Pattern Recognition},
  year      = {2025}
}

@article{liu2025s2edit,
  title   = {S$^2$Edit: Text-Guided Image Editing with Precise Semantic and Spatial Control},
  author  = {Liu, Xudong and Chen, Zikun and Jiang, Ruowei and Wu, Ziyi and Yin, Kejia and Zhao, Han and Aarabi, Parham and Gilitschenski, Igor},
  journal = {arXiv preprint arXiv:2507.04584},
  year    = {2025}
}

@inproceedings{bai2024edicho,
  title     = {Edicho: Consistent Image Editing in the Wild},
  author    = {Bai, Qingyan and Ouyang, Hao and Xu, Yinghao and Wang, Qiuyu and Yang, Ceyuan and Cheng, Ka Leong and Shen, Yujun and Chen, Qifeng},
  booktitle = {Proceedings of the IEEE/CVF International Conference on Computer Vision},
  year      = {2025}
}

@inproceedings{zhang2018perceptual,
  title     = {The Unreasonable Effectiveness of Deep Features as a Perceptual Metric},
  author    = {Zhang, Richard and Isola, Phillip and Efros, Alexei A. and Shechtman, Eli and Wang, Oliver},
  booktitle = {Proceedings of the IEEE Conference on Computer Vision and Pattern Recognition},
  year      = {2018}
}

@article{wang2004ssim,
  title   = {Image Quality Assessment: From Error Visibility to Structural Similarity},
  author  = {Wang, Zhou and Bovik, Alan C. and Sheikh, Hamid R. and Simoncelli, Eero P.},
  journal = {IEEE Transactions on Image Processing},
  volume  = {13},
  number  = {4},
  pages   = {600--612},
  year    = {2004}
}

@inproceedings{radford2021clip,
  title     = {Learning Transferable Visual Models from Natural Language Supervision},
  author    = {Radford, Alec and Kim, Jong Wook and Hallacy, Chris and Ramesh, Aditya and Goh, Gabriel and Agarwal, Sandhini and Sastry, Girish and Askell, Amanda and Mishkin, Pamela and Clark, Jack and Krueger, Gretchen and Sutskever, Ilya},
  booktitle = {Proceedings of the International Conference on Machine Learning},
  year      = {2021}
}

@inproceedings{liu2024groundingdino,
  title     = {Grounding {DINO}: Marrying {DINO} with Grounded Pre-training for Open-Set Object Detection},
  author    = {Liu, Shilong and Zeng, Zhaoyang and Ren, Tianhe and Li, Feng and Zhang, Hao and Yang, Jie and Jiang, Qing and Li, Chunyuan and Yang, Jianwei and Su, Hang and Zhu, Jun and Zhang, Lei},
  booktitle = {Proceedings of the European Conference on Computer Vision},
  year      = {2024}
}

@inproceedings{kirillov2023sam,
  title     = {Segment Anything},
  author    = {Kirillov, Alexander and Mintun, Eric and Ravi, Nikhila and Mao, Hanzi and Rolland, Chloe and Gustafson, Laura and Xiao, Tete and Whitehead, Spencer and Berg, Alexander C. and Lo, Wan-Yen and Doll{\'a}r, Piotr and Girshick, Ross},
  booktitle = {Proceedings of the IEEE/CVF International Conference on Computer Vision},
  year      = {2023}
}

@misc{beaudouin2025drfs,
  title         = {Delta Rectified Flow Sampling for Text-to-Image Editing},
  author        = {Beaudouin, Gaspard and Li, Minghan and Kim, Jaeyeon and Yoon, Sung-Hoon and Wang, Mengyu},
  year          = {2025},
  eprint        = {2509.05342},
  archiveprefix = {arXiv},
  primaryclass  = {cs.CV}
}

@misc{luo2026vags,
  author        = {Luo, Yan and Aidara, Ahmadou and Lu, Jingyi and Moebel, Jeremy and Han, Kai and Wang, Mengyu},
  title         = {{VAGS}: Velocity Adaptive Guidance Scale for Image Editing and Generation},
  year          = {2026},
  eprint        = {2605.15661},
  archiveprefix = {arXiv},
  primaryclass  = {cs.CV}
}
